\documentclass{article}
\usepackage[utf8]{inputenc}
\usepackage{amsmath}
\usepackage{amssymb}
\usepackage{mathrsfs}
\usepackage{graphicx}
\usepackage{soul}
\usepackage{color}
\usepackage{multirow}
\usepackage{booktabs} 
\usepackage{geometry}
\usepackage{subfigure}
\usepackage[breaklinks,colorlinks,linkcolor=black,citecolor=black,urlcolor=black]{hyperref}
 \usepackage{graphicx}
 \usepackage{authblk}
\newcommand*{\blue}{\textcolor{black}}

\title{Reachability-Based Confidence-Aware Probabilistic  Collision  Detection in Highway Driving}

\author[1,2]{Xinwei Wang\thanks{Corresponding author. This research was supported by the SAFE-UP project (proactive SAFEty systems and tools for a constantly UPgrading road environment). SAFE-UP has received funding from the European Union’s Horizon 2020 research and innovation programme under Grant Agreement 861570.  Email: xinwei.wang@qmul.ac.uk, z.li@bit.edu.cn, j.alonsomora@tudelft.nl, meng.wang@tu-dresden.de }}
\author[3,4]{Zirui Li}
\author[2]{Javier Alonso-Mora}
\author[4]{Meng Wang}

\affil[1]{School of Engineering and Materials Science,
Queen Mary University of London, E1 4NS London, UK}
\affil[2]{Department of Cognitive Robotics,
Delft University of Technology, 2628 CD Delft, The Netherlands}
\affil[3]{School of Mechanical Engineering, Beijing Institute of Technology, 100081 Beijing, China}
\affil[4]{Chair of Traffic Process Automation, "Friedrich List" Faculty of Transport and Traffic Sciences, TU Dresden, 01069 Dresden, Germany}

\begin{document}

\maketitle

{\bf Abstract:} Risk assessment is a crucial component of collision warning and avoidance systems in intelligent vehicles. To accurately detect potential vehicle collisions, 
reachability-based formal approaches have been developed to ensure driving safety, but suffer from over-conservatism, potentially leading to false-positive risk events in complicated real-world applications. In this work, we combine two reachability analysis techniques, i.e., backward reachable set (BRS) and stochastic forward reachable set (FRS), and propose an integrated probabilistic collision detection framework in highway driving. Within the framework, we can firstly use a BRS to formally check whether a two-vehicle interaction is safe; otherwise, a prediction-based stochastic FRS is employed to estimate a collision probability at each future time step. In doing so, the framework can not only  
identify non-risky events with guaranteed safety, but also provide accurate collision risk estimation in safety-critical events. To construct the stochastic FRS, we develop a neural network-based  acceleration model for surrounding vehicles, and further incorporate  confidence-aware dynamic belief  to improve the prediction accuracy. 
Extensive experiments are conducted to validate the performance of the acceleration prediction model based on naturalistic highway driving data, and the efficiency and effectiveness of the framework with the infused confidence belief are tested both in naturalistic and simulated highway scenarios. The proposed risk assessment framework is promising in real-world applications.

\section{Introduction}

Autonomous vehicles (AVs) are expected to significantly benefit future  mobility, while one of the prerequisites for enabling AVs publicly available is to ensure autonomous driving safety~\cite{kalra2016driving}. Highways are structured environments designed for vehicles to drive at a consistently high speed for efficient road trips, and are the first applications of Level 1 and Level 2 automated vehicles.  In the transition from human-driven and lower-level automated vehicles to high-level AVs, it is essential to address driving safety on highways both for conventional vehicles and AVs.  
To identify driving risk and potential vehicle crashes, extensive research on risk assessment and collision detection has been conducted~\cite{lefevre2014survey,mukhtar_vehicle_2015}. 
To accurately detect potential vehicle collisions, reachability-based formal approaches have been developed~\cite{pek2020using}, since they can mathematically check whether the behavior of a system,  satisfies given safety requirements.

Reachability analysis (RA) has been widely employed to formally verify driving safety~\cite{falcone2011predictive,althoff2014online}. RA computes a complete set of states that an agent (e.g.\@ a vehicle) can reach given an initial condition within a certain time interval~\cite{althoff2021set}. Based on RA, a safety verification thus can be performed by propagating all possible reachable space of the AV and other traffic participants on the road. In doing so, safety is ensured if such forward reachable set (FRS) of the automated vehicle does not intersect that of other traffic participants during the propagation period. In line with such a definition, the FRS can formally verify safety between road users, but easily lead to over-conservative results because the state propagation is feedforward and ignore traffic participant interactions (i.e., vehicles react to the surrounding environment and adjust the control output)~\cite{althoff2021set}.

Alternatively, RA can be conducted in a closed-loop manner~\cite{leung2020infusing}. Given a target set representing a set of undesirable states (e.g., collision states between two vehicles) and worst-case disturbances, we define the backward reachable set (BRS) as the set of states that could lead to being in the target set during a certain time horizon. Specifically, a BRS is the state set in which a control strategy does not exist  to prevent the AV from the target set under worst-case disturbances. An unsafe area thus can be directly identified by the BRS with initial vehicle states. Note that one can compute BRS offline in advance, and then use the cached BRS in real-time. Although BRS considers control reactions from the AV and is less conservative compared to FRS, it still suffers from over-conservatism due to the worst-disturbance closed-loop reactions. 
 
We aim to use the RA for driving risk evaluation and potential collision detection. However, both these two RA approaches suffer from over-conservatism. To reduce the over-conservative nature of forward reachability, the time horizon for an FRS is typically kept small and is recomputed frequently. Although BRS incorporates a closed-loop feedback to consider the worst disturbance from the surrounding vehicle, general interactions between vehicles are not a pursuit-evasion~\cite{chapman2021risk}. It is reasonable to consider a more realistic situation: \textit{the interactions are not adversarial, but leading to crashes is still possible}.

\subsection{Related work} \label{sec:review}
Driving risk assessment is crucial to identify potential collision and quantify risk level. Various Surrogate Measures of Safety (SMoS) have been constructed to \blue{measure driving risk}. Typically, SMoS can be calculated in a time-series manner, including Time To Collision~\cite{lee1976theory,minderhoud2001extended}, Time Headway~\cite{vogel2003comparison}, and Time to Lane Crossing~\cite{mammar2006time}. Unfortunately, these developed SMoS are mostly deterministic, which means that uncertainties  in vehicle motion and environment are not considered. \blue{Although several probabilistic approaches~\cite{saunier2008probabilistic,davis2011outline,kuang2015tree,mullakkal2020probabilistic} have been integrated to improve the performance of SMoS, these methods could  suffer from the additional computational load (especially for long-term prediction) and cannot formally ensure driving risk.} 

One could also assess driving risk by estimating the {current and future} collision probability given surrounding road participants. Collision detection can be generally divided into three methodologies, i.e., neural network-based approaches, probabilistic approaches, and formal verification approaches. 
Neural networks have the potential to provide accurate vehicle collision detection by classifying safety-critical scenarios. For instance, a collision detection model using a neural network-based classifier was developed in~\cite{katare_embedded_2019}. The proposed model takes onboard sensor data, including acceleration, velocity, and separation distance, as input to a neural network based classifier, and outputs whether alerts are activated for a possible collision. A specific animal  detection approach was proposed in~\cite{saleh_kangaroo_2016}, where a deep semantic segmentation convolutional neural network is trained to recognize and detect animals in dynamic environments. Although neural network-based approaches are effective to identify potential collisions, the trained classifier generally cannot include clear decision rules and is hard to interpret. 
 
\blue{To address uncertainties of surrounding road participants and better estimate collision probability, prediction based approaches have also been widely adopted.} A conceptual framework to analyze and interpret the dynamic traffic scenes was designed in~\cite{laugier_probabilistic_2011} for collision estimation. The collision risks are estimated as stochastic variables and predicted  relying on driver behavior evaluation with hidden Markov models. \blue{To calculate the collision probability, an intention estimation and a long-term trajectory prediction module are combined to construct a probability field for future vehicle positions~\cite{annell_probabilistic_2016}}. Given a set of local path candidates, a collision risk assessment considering lane-based probabilistic motion prediction of surrounding vehicles was proposed in~\cite{kim_collision_2018}. However, these methods typically require pre-defined parameters of position distributions, which can  impact the adaptability of the probabilistic collision detection.

Formal verification approaches, which can formally ensure system safety given specific control input range and safety requirements, have also been employed to address collision detection~\cite{pek2017verifying,tornblom2019abstraction}. As one of the formal approaches, RA computes a complete set of states that an agent (e.g. a vehicle) can reach given an initial condition within a certain time interval~\cite{althoff2021set}. Based on RA, a safety verification thus is performed by propagating all possible reachable space of the AV and other traffic participants forward in time and checking the overlaps.  To reduce the over-conservative nature of  forward reachability, a stochastic FRS discretizing the reachable space into grids with probability distributions was developed~\cite{althoff2009model}. At each time step, a collision probability is provided by summing probabilities of the states that vehicles 
intersect. However, this approach is based on Markov chains, which assume that the vehicle state and its control input evolve only in line with  the current state. Besides, it cannot explicitly address two-dimensional motion, as lane-change maneuvers are not considered.

RA can also be conducted in a closed-loop manner with worst-case disturbances, namely the BRS~\cite{leung2020infusing}.  Although BRS can be constructed offline by employing advanced \blue{Hamilton-Jacobi-Isaacs partial differential equation (HJI PDE)} solvers~\cite{bansal2017hamilton}, it  suffers from over-conservatism due to the worst-disturbance assumptions, which are not realistic. To fill the research gap, we aim to combine the BRS and stochastic FRS technique into an integrated collision detection framework, which cannot only theoretically ensure safety in non-risky interactions, but also provide an accurate collision estimation in safety-critical scenarios.

 \subsection{Objectives and contribution}

\begin{figure}[htb]
	\begin{center}		\includegraphics[width=0.7\textwidth]{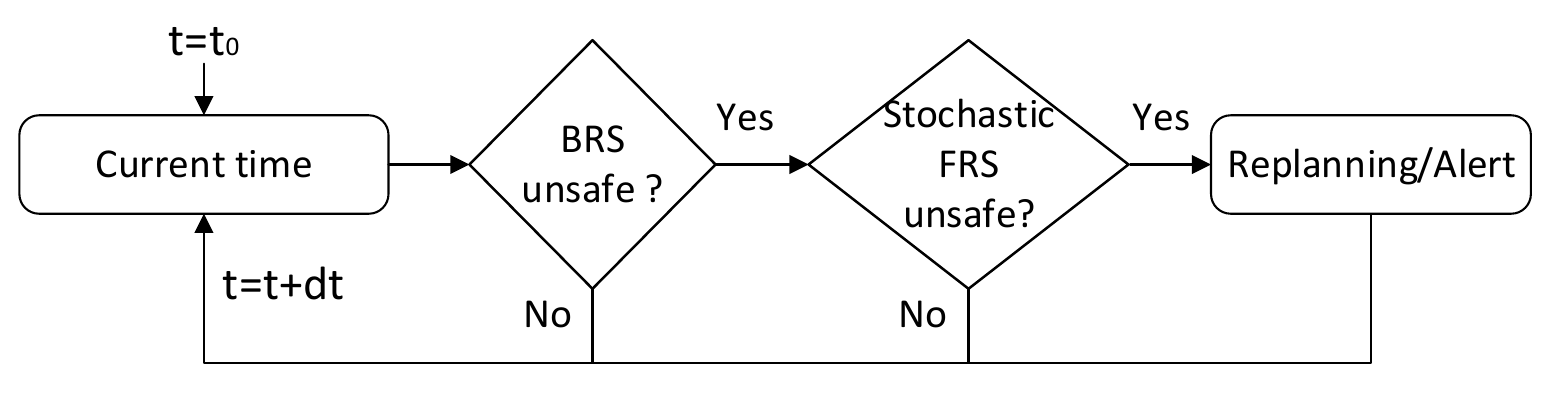}
		\caption{Brief diagram of the integrated collision detection framework.
}\label{fig:framework_ini}
	\end{center}
\end{figure}

\blue{The stochastic FRS is capable of accurate collision estimation when integrated with a prediction model. However, it cannot provide a safety guarantee, as the employed probabilistic distribution predictor could  fail under corner cases, resulting in false negative/positive cases (e.g., in a safe cut-in scenario, BRS can ensure safety while stochastic FRS estimates a biased collision probability, resulting in a false positive alarm). This motivates us to propose an integrated collision detection framework including both BRS and stochastic FRS to evaluate highway driving risk}; see Fig.~\ref{fig:framework_ini}. The BRS is firstly computed based on HJI PDE~\cite{leung2020infusing}; if the relative positions of vehicles are identified unsafe by the BRS, a 
stochastic FRS considering surrounding vehicle manoeuvring modes is further established to calculate a collision probability at each future time step. Here the stochastic FRS shares the same reachable states as FRS. In addition, each state of a stochastic FRS has an estimated probability. It is ideal to directly use a stochastic BRS for collision detection, while the computation of a stochastic BRS is not readily viable due to the closed-loop form of BRS. 

Based on the stochastic FRS, a collision probability between two vehicles can be calculated by summing up state probabilities where two vehicles spatially overlap. If the obtained collision probability is above a pre-defined threshold, the ego vehicle has to execute an emergency brake or swerve to avoid crashes with the other vehicle. The proposed framework benefits from both BRS and FRS: the driving safety could be theoretically ensured when the relative vehicle positions are out of the unsafe area identified by the BRS; otherwise the framework provides a collision probability based on a developed stochastic FRS. 

To construct the stochastic FRS,  we develop a long-short term memory (LSTM) model for multi-maneuver acceleration prediction on highways. The proposed model has two stages for maneuver prediction (i.e., lane-keeping, turning-left/-right on highways) and acceleration prediction respectively, and the  model input features are also selected differently at each stage.  We further incorporate a confidence-aware belief vector to generate a group of predicted acceleration distributions, which can dynamically adjust the degree of confidence inferred from  current prediction accuracy~\cite{fridovich2020confidence}. The confidence-aware belief vector could result in a concentrated stochastic FRS when the LSTM model has higher prediction accuracy, and lead to a more spread stochastic FRS when vehicles move  unexpectedly.

The main contribution in this work is summarized as:
  \begin{itemize}
  \item  We propose a multi-modal acceleration prediction model for surrounding vehicles, and establish stochastic FRS for each surrounding vehicle by leveraging the proposed acceleration predictor. Furthermore, we incorporate confidence awareness to generate a group of predicted acceleration distributions, and  dynamically update the degree of confidence, leading to a more accurate stochastic FRS and more agile collision detection results.
    \item An integrated probabilistic collision detection framework including both BRS and stochastic FRS is proposed to evaluate the highway driving risk. Within the framework, an  offline-computed and cached BRS is used online to check whether the car-car interaction safety can be theoretically ensured; if not,  a stochastic FRS is then computed online to provide an accurate collision probability at each future time step. 
 \end{itemize}

 We have presented the results of stochastic FRS using the LSTM prediction model in~\cite{9827304}. In this paper, we  significantly extend~\cite{9827304} by including the integrated collision detection framework  and infusing a confidence-aware belief vector for more accurate stochastic FRS. Extensive and comprehensive experiments, which are different from those in~\cite{9827304},  have been conducted to validate  the proposed framework. The remainder of the paper is as follows: 
 Section~\ref{sec:preli} provides preliminaries on BRS, FRS and Markov-based stochastic FRS. In Section~\ref{sec:models}, a specific prediction-based confidence-aware stochastic FRS is developed, and we establish an integrated driving risk framework including BRS and stochastic FRS in Section~\ref{sec:frame}. Extensive experiments are conducted in Section~\ref{sec:sim} and we conclude our work in Section~\ref{sec:conc}.

\section{Preliminaries} \label{sec:preli}

\subsection{Backward reachability set (BRS)}
 Backward reachability analysis is regarded as an optimal control problem and thus computing the reachable set is equivalent to solving the HJI PDE~\cite{leung2020infusing}. Define the system dynamics by $\dot{\mathbf{x}} = f (\mathbf{x}, \mathbf{u}, \mathbf{d})$ where ${\mathbf{x}} \in \mathbb{R}^{n_1} $ and $\mathbf{u} \in \mathcal{U} \subset \mathbb{R}^{n_2}$ are the state and control \blue{input ($\mathcal{U}$ is the admissible control input set)},  $\mathbf{d} \subset \mathbb{R}^{n_3}$ is the disturbance \blue{and $n_{i} \in \mathbb{Z}^+ (i=1,2,3)$ are the dimensions}. The system dynamics are assumed to be uniformly continuous and bounded. 
 
 In the context of two-vehicle interactions, $\mathbf{u}$ corresponds to the control of ego vehicle, and $\mathbf{d}$ corresponds to the control of the surrounding vehicle, since its actions are treated as disturbance inputs. Specifically, let $(\mathbf{x}_{e}, \mathbf{u}_{e})$/$(\mathbf{x}_{s}, \mathbf{u}_{s})$ represent the state and control of ego/surrounding vehicle, and $\mathbf{x}_{rel}$ be the system states between vehicles, e.g., relative two-dimensional distance $(y_{1,rel},y_{2,rel})$ and velocity $(v_{1,rel},v_{2,rel})$. 
 Thus the system dynamics are given by $\dot{\mathbf{x}}_{ref} = f (\mathbf{x}_{ref}, \mathbf{u}_{e}, \mathbf{u}_{s})$. The formal definition of the BRS, denoted by $\mathcal{BR}(t)$, for the relative system is
\begin{equation}
\begin{split}
    \mathcal{BR}(t) := \{ \bar{\mathbf{x}}_{rel}  : \exists  \mathbf{u}_{s}(\cdot), \forall \mathbf{u}_{e}(\cdot),\exists  t^{\star} \in [t,0],\\ \mathbf{x}_{rel}(t) = \bar{\mathbf{x}}_{rel} \wedge \dot{\mathbf{x}}_{rel} = f(\mathbf{x}_{rel}, \mathbf{u}_{e}, \mathbf{u}_{s}) \wedge \mathbf{x}_{rel}(t^{\star})\in \mathcal{T} 
    \}
\end{split}
\end{equation}
Here $\mathcal{BR}(t)$ represents the set of unsafe states $\bar{\mathbf{x}}_{rel}$ at time $t$, from which if the surrounding vehicle followed an adversarial policy $\mathbf{u}_{s}$, any policy $\mathbf{u}_{e}$ would lead to the state set $\mathcal{T}$  \blue{at $t^{\star} \in [t,0]$} where two vehicles collide within a time horizon. 

Assuming optimal (i.e., adversarial) surrounding vehicle actions, $\mathcal{BR}(t)$ can be computed by defining a value function $V (t, \mathbf{x}_{rel})$ which obeys the HJI PDE, where the solution $V (t, \mathbf{x}_{rel})$ gives the BRS as its zero sublevel set:
\begin{equation}
    \mathcal{BR}(t) = \{\mathbf{x}_{rel} : V (t, \mathbf{x}_{rel}) \leq 0 \}
\end{equation}
The HJI PDE is solved starting from the boundary condition $V (t, \mathbf{x}_{rel})$, which indicates whether  state $\mathbf{x}_{rel}$ belongs to collision set $\mathcal{T}$. We cache the solution $V (t, \mathbf{x}_{rel})$ to be used online as a look-up table.

\subsection{Forward reachability set (FRS) and Markov-based stochastic FRS}



The computation of an FRS is done by considering all possible control inputs $\mathbf{u}\in \mathcal{U}$ of a system $\dot{\mathbf{x}} = f (\mathbf{x}, \mathbf{u})$ given an initial set of states $\mathcal{X}_{0}$. 
The FRS of a system is formally defined as


\begin{equation}
    \mathcal{FR}(t) := \{ \bar{\mathbf{x}}: \exists \mathbf{u}(\cdot), \exists t^{\star} \in [0,t], \mathbf{x}(0)\in \mathcal{X}_{0} \wedge \dot{\mathbf{x}} = f (\mathbf{x}, \mathbf{u}, \mathbf{d})
    \wedge  \mathbf{x}(t^{\star}) = \bar{\mathbf{x}}\}
\end{equation}
where $\mathcal{FR}(t)$ is a forward reachable set of states $\bar{\mathbf{x}}$  at time \blue{$t^{\star} \in [0,t]$} from an initial state $\mathbf{x}(0)\in \mathcal{X}_{0}$ at the current time $0$ and subject to any input~$\mathbf{u}$ belonging to the admissible control input set~$\mathcal{U}$.

Based on the definition of FRS, we can further mathematically formulate a stochastic version of FRS as $\mathcal{SFR}(t)$, where each state $\mathbf{\bar{x}}$ within the FRS is associated with a state probability $p(\bar{\mathbf{x}})$.
\begin{equation}
    \mathcal{SFR}(t) := \{ (\bar{\mathbf{x}},p(\bar{\mathbf{x}})): \exists \mathbf{u}(\cdot), \exists t^{\star} \in [0,t], \mathbf{x}(0)\in \mathcal{X}_{0} \wedge \dot{\mathbf{x}} = f (\mathbf{x}, \mathbf{u}, \mathbf{d})
    \wedge  \mathbf{x}(t^{\star}) = \bar{\mathbf{x}}\}
\end{equation}

One of the most frequently used techniques is to approximate stochastic processes by Markov chains, 
 which present a stochastic dynamic system with discrete states
 ~\cite{althoff2009model}.  The discretized time step series are denoted as $\{0,1,\dots,e\}$, where $e$ is the future final time step, and the duration of each time step is $dt$. Due to the stochastic characteristics, the system state at the predicted time step is not exactly known, and a probability $p_{i}({k})$ is assigned to each state $i$ at the  time step ${k}$ (the discretized system state is denoted as $i,j$ for simplicity). Then the probability vector $\mathbf{p}({k+1})$ composed of probabilities $p_{i}({k})$ over all states is updated as
 

\begin{equation}
   \mathbf{p}({k+1})=\Phi \cdot \mathbf{p}({k}) \label{pTrans}
\end{equation}
where $\Phi$ is the state transition matrix. Here $\Phi$ is time invariant as the model is assumed as Markovian.

To implement a Markov chain model, the system state first needs to be discretized if the original system is continuous. For the vehicle dynamic system, we represent it as a tuple with four discretized elements, including two-dimensional vehicle positions and velocities. Meanwhile, the control input requires to be discretized. Detailed discretization parameters are reported in Section~\ref{sec:setup}.

Each element $\Phi_{ji}$ in matrix $\Phi$ represents the state transition probability from state $i$ to $j$. 
Note that the transition probabilities depend on the discrete input $\mathbf{u}$ as well, i.e., each discrete input $\mathbf{u}$ generates a conditional transition probability matrix $\Phi^{u}$. Specifically, each element $\Phi^{u}_{ji}$ in the conditional matrix $\Phi^{u}$ is the possibility starting from the initial state $i$ to $j$ under  control $\mathbf{u} \in \mathcal{U}$, where  $\mathbf{u}$ represents the corresponding control input of $\Phi_{ji}^{u}$. The conditional probability $\Phi_{ji}^{u}$ therefore is expressed as 


\begin{equation} \label{eq:phi_u}
\Phi_{ji}^{u} =
\left\{
\begin{aligned}
        p_{i}^{u}& ,~\text{if state} \ {i} \ \text{reaches state} \ j \ \text{with input}\ \mathbf{u}\\
        0& ,~\text{otherwise}
        \end{aligned}
        \right.
\end{equation}
where $p^{u}_{i}$ is the control input probability given state $i$. The time index does not appear here as it is a Markov process. The overall state transition matrix is then constructed as

\begin{equation}  \label{eq:phi}
   \Phi_{ji} = \sum\limits_{u \in\mathcal{U} } \Phi_{ji}^{u}
\end{equation}

The probability distribution of the control input $p^{u}_{i}$ is dynamically changed by another Markov chain with transition matrix $\Gamma_i$, depending on the system state $i$. This allows a more accurate modeling of driver behavior by considering the frequency and intensity of the changes of control input. As a consequence, the transition matrices $\Gamma$ have to be learned by observation or set by a combination of simulations and heuristics. By incorporating the two transition matrices $\Phi$ and $\Gamma$, a Markov-based stochastic FRS with probabilities $\mathbf{p}({k})$ over all discretized states can be obtained at each predicted time step $k$.  

In~\cite{althoff2009model}, the  control  transition probability matrices $\Gamma$ only depend on the current control input and the state at the current time.  While the computational efficiency is ensured by using such simplified Markovian setting, the future control input and trajectories of a vehicle can be influenced by historical information and interactions with surrounding environment~\cite{deo2018convolutional}. Therefore in this work, we do not assume that the vehicle system state is Markovian. Instead, to address the historical information and vehicle interactions, we aim to use a vehicle control predictor with multi-maneuvering modes to generate and dynamically update the transition matrices at each time step $k$ as

\begin{equation}  \label{eq:phi_dy}
   \Phi_{ji}(k) = \sum\limits_{u \in\mathcal{U} } \Phi_{ji}^{u}(k)
\end{equation}

 \section{Prediction-based confidence-aware stochastic FRS}\label{sec:models}
 
 In this section, to provide more accurate prediction of surrounding vehicles, we first introduce a two-stage multi-modal acceleration prediction model consisting of a lane change maneuver prediction module and an acceleration prediction module. Then we detail how the stochastic FRS is established by incorporating the proposed acceleration prediction model and infusing a confidence belief vector. Section~\ref{sec:4.1} and Section~\ref{sec:4.2} have been presented in~\cite{9827304}, and are included here for completeness. Note that the prediction model is replaceable, as long as the accelerations can be predicted by bi-variate normal distributions.

 \subsection{Acceleration prediction of a surrounding vehicle}\label{sec:4.1}
 
   \begin{figure*}[!tb]
\label{sec:model}
	\begin{center}
		\includegraphics[width=0.9\textwidth]{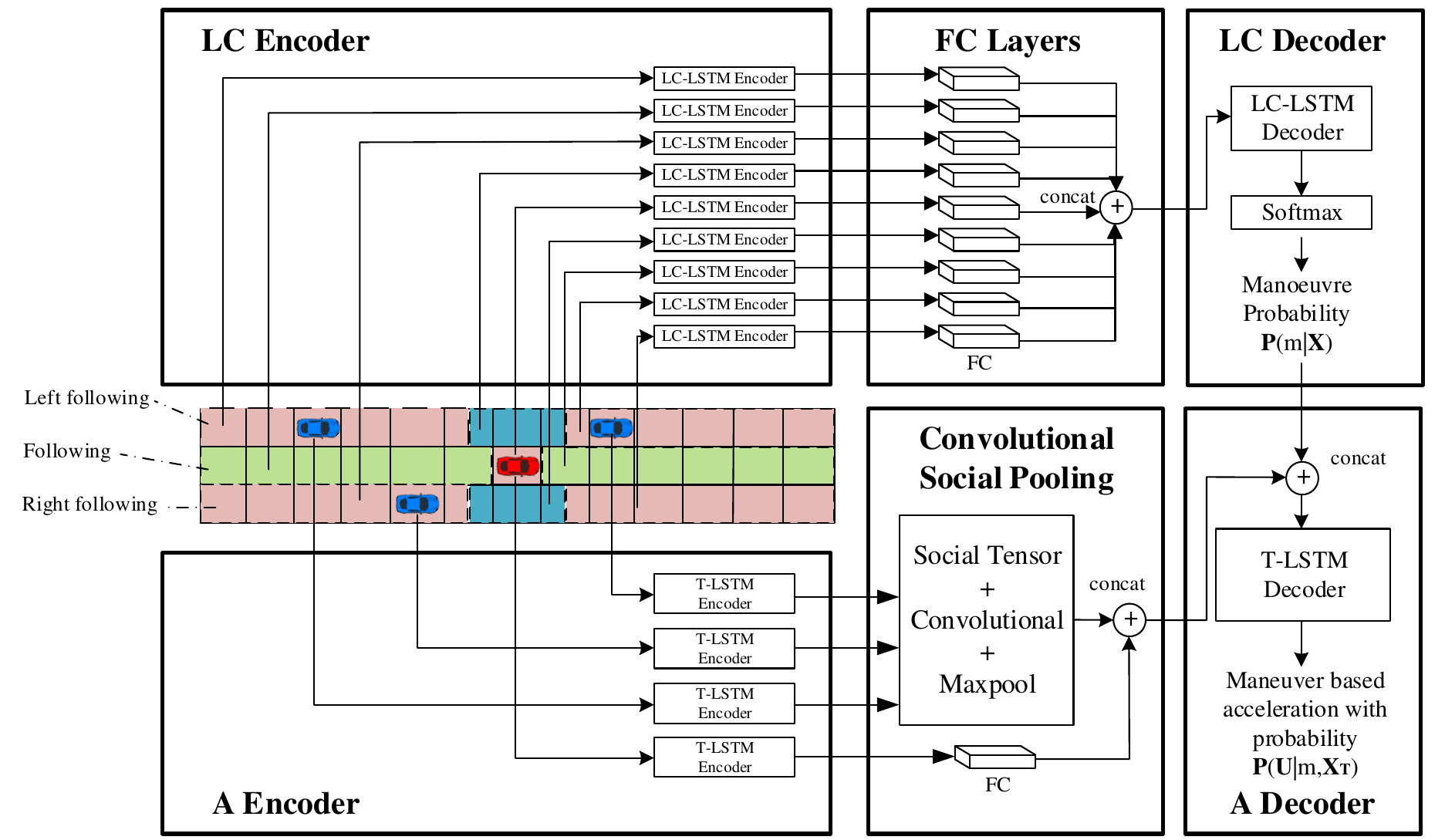}
		\caption{Overview of the acceleration prediction model,  consisting of a lane-change maneuver prediction module and an acceleration  prediction module (denoted as LC and A respectively in the figure). The two modules both have an encoder-decoder structure, but adopt and process historical information as input in different ways. Abbreviations concat and FC stand for the concatenation operation and fully connect layer respectively. A
variant of the model for trajectory prediction was developed earlier in~\cite{wang2021probabilistic}.  }\label{fig:model}
		\end{center}
		\end{figure*}
		
 Typically, the vehicle trajectories and acceleration are predicted using both current and historical information~\cite{deo2018convolutional}. In doing so, the prediction accuracy can be improved compared to that only using  current states as input. This motivates us to establish an LSTM based network to  predict probabilistic future vehicle accelerations both using current and historic vehicle information. An overview of the developed two-stage acceleration prediction model is illustrated in Fig.~\ref{fig:model}.
 
 
 \subsubsection{Two-stage vehicle acceleration prediction}
We have developed a two-stage multi-modal trajectory prediction model in~\cite{wang2021probabilistic}. In this work,  we keep the same lane-change maneuver prediction model at the first stage, but develop a new acceleration prediction model at  the second stage. This is necessary because the acceleration prediction is employed to enable the dynamic update of the conditional probability $\Phi_{ji}^{u}(k)$ in Eq.~\eqref{eq:phi_dy}.

We first briefly introduce the adopted lane-change maneuver prediction module from~\cite{wang2021probabilistic}. The input of the module is expressed as 

\begin{equation}
\mathbf{X} = [\mathbf{x}^{(-h)},\dots,\mathbf{x}^{(-1)},\mathbf{x}^{(0)}]
\end{equation}
where $\mathbf{X}$ represents all input features from previous time step $-h$ to the current time step $0$. At each historic time step, the collected input is composed of three parts: $\mathbf{x}^{(\cdot)} = [\mathbf{x_{T}}^{(\cdot)},\mathbf{b}^{(\cdot)},d^{(\cdot)}]$, where $\mathbf{x_T}^{(\cdot)}$ is the position information for vehicle being predicted as well as its surrounding vehicles, $\mathbf{b}^{(\cdot)}$ contains two binary values to check whether the predicted vehicle can turn left and right, and $d^{(\cdot)} \in [-1,1]$ is the normalized deviation value from the current lane center.

As shown on the top of Fig.~\ref{fig:model}, LSTMs are used to encode and decode the lane-change maneuver prediction model, in which the encoding information is passed to fully connected layers before decoding. The output of the model is a probability distribution $\mathbf{P}(m|\mathbf{X})$ for each lane-change maneuver mode m (i.e. change to left, change to right, keep the same lane) time step $1$ to $e$.

As for the acceleration prediction at the second stage, the input includes historic positions of the vehicle being predicted and surrounding vehicles, in addition to the historic accelerations $\mathbf{x_A}^{(-h:0)}$ of the vehicle being predicted:
\begin{equation}
\mathbf{X_{T}} = [\mathbf{x_T}^{(-h)},\dots,\mathbf{x_T}^{(-1)},\mathbf{x_T}^{(0)}, \mathbf{x_A}^{(-h:0)}]
\end{equation}

As we use additional acceleration information for the vehicle being predicted, we modify the input size of the LSTM encoder in~\cite{wang2021probabilistic} for the vehicle being predicted, while maintaining the overall network structure unchanged. Detailed information of the second-stage model is referred to~\cite{deo2018convolutional,wang2021probabilistic}.

Given the input $\mathbf{X_T}$ and corresponding maneuver mode probability distribution $\mathbf{P}(m|\mathbf{X})$, the output $\mathbf{P}(\mathbf{U}|m,\mathbf{X_T})$ of the second-stage acceleration prediction model is conditional acceleration distributions over 
\begin{equation}
\mathbf{U} = [\mathbf{u}^{({1})},\dots,\mathbf{u}^{(e)}]
\end{equation}
where $\mathbf{u}^{(\cdot)}$ is the predicted vehicle acceleration at each time step within the prediction horizon. 
Note that the prediction horizon and time increment are the same as those for the reachable set computation, respectively.

Given the three defined maneuvers $m$, the probabilistic multi-modal distributions are calculated as 
\begin{equation}
\mathbf{P}(\mathbf{U}|\mathbf{X})  = \sum_{m} \mathbf{P}_\Theta (\mathbf{U}|m,\mathbf{X_T})\mathbf{P}(m|\mathbf{X})
\end{equation}
where outputs $\Theta = [\Theta^{({1})},\dots,\Theta^{(e)}]$ are time-series bivariate normal distributions.  $\Theta^{(k)} = \cup_{m}\{\mu_{1m}^{k},\mu_{2m}^{k},\sigma_{1m}^{k},\sigma_{2m}^{k},\rho_{m}^{k}\}$ corresponds to the predicted acceleration means and standard deviations along two dimensions, and the correlation at  future  time  step $k$ under each maneuver mode $m$, respectively. 

Under acceleration distributions $\Theta$, the future vehicle trajectories are propagated as 
\begin{equation}\label{eq:propagate}
\left \{
\begin{aligned}{}
    v_{1m}^{k+1} =  &  v_{1m}^{k} + \mu_{1m}^{k+1}dt \\
    v_{2m}^{k+1} =  &  v_{2m}^{k} + \mu_{2m}^{k+1}dt\\
    y_{1m}^{k+1} =  &   y_{1m}^{k}  + (v_{1m}^{k+1} + v_{1m}^{k})dt/2 \\
    y_{2m}^{k+1} =  &   y_{2m}^{k}  + (v_{2m}^{k+1} + v_{2m}^{k})dt/2
\end{aligned}
\right.
\end{equation} 
where $dt$ is the time increment, $v_{1m}^{k}, v_{2m}^{k}, y_{1m}^{k},y_{2m}^{k}$ are the propagated two-dimensional velocities and positions at  future  time  step $k$ for each maneuver mode $m$, respectively. $(v_{1m}^{0}, v_{2m}^{0}, y_{1m}^{0},y_{2m}^{0})$ denotes the system state at the current time $t$. 

\blue{In order to properly propagate probabilistic distributions  from  accelerations to trajectories,  we assume that the  trajectory variance and correlation propagation only depend on the acceleration at the current time, rather than previous time steps. Otherwise, the trajectory variance/correlation at the future $k$ step is to be determined by the acceleration variances/correlations from the current time step to the future $k$ step, which makes the probabilistic distribution propagation complicated and intractable. The propagated trajectory standard deviations can then be obtained as $\widetilde{\sigma}_{1m}^{k} = {\sigma}_{1m}^{k}\cdot(dt)^{2}/2$ and $\widetilde{\sigma}_{2m}^{k} = {\sigma}_{2m}^{k}\cdot(dt)^{2}/2$, and the correlation  as $\rho_{m}^{k}\cdot(dt)^{2}/2$.}
Therefore, the propagated probabilistic distributions  of the vehicle position are expressed as $\widetilde{\Theta}^{(k)} =\{y_{1m}^{k},y_{2m}^{k},\widetilde\sigma_{1m}^{k},\widetilde\sigma_{2m}^{k},\rho_{m}^{k}\}_{m=\{1,2,3\}}$\@.

 \subsubsection{Model training}
 Typically a multi-modal prediction model is trained to minimize the negative log likelihood (NLL) of its conditional distributions as
 
\begin{equation} \label{eq:nll}
-\text{log}\left(\sum_{m} \mathbf{P}_\Theta (\mathbf{U}|m,\mathbf{X_{T}})\mathbf{P}(m|\mathbf{X})\right)
\end{equation}
For more accurate collision probability estimation,  we focus on the potential  collision when two vehicles have intersections along the trajectories. We therefore directly minimize the trajectory prediction errors propagated from the acceleration prediction in line with~\cite{zhou2017recurrent} as 
\begin{equation} \label{eq:nll-revised}
-\text{log}\left(\sum_{m} \mathbf{P}_{\widetilde{\Theta}} (\mathbf{Y}|m,\mathbf{X_{T}})\mathbf{P}(m|\mathbf{X})\right)
\end{equation}
where $\mathbf{Y}= [\mathbf{y}^{(1)},\dots,\mathbf{y}^{(f)}]$ is the propagated trajectories with distributions $\widetilde{\Theta}$, and $\mathbf{y}^{(k)} = \{y_{1m}^{k},y_{2m}^{k}\}$ are the predicted positions of the vehicle at time step $k$ under maneuver mode $m$.

To further improve the prediction performance, we separately train the lane-change maneuver and vehicle acceleration prediction models. This is because that the proposed approach has a two-stage structure: the maneuver probabilities are first predicted, and then for the corresponding conditional vehicle acceleration distributions.   For the maneuver prediction model, it is trained to minimize the NLL  of the  maneuver probabilities $-\text{log}\left(\sum_{m} \mathbf{P}(m|\mathbf{X})\right)$; for the vehicle acceleration prediction, the adopted model is to minimize $-\text{log}\left( \sum_{m} \mathbf{P}_{\widetilde{\Theta}} (\mathbf{Y}|m,\mathbf{X_T})  \right)$.
 
 \subsection{Prediction-based stochastic FRS of a surrounding vehicle}\label{sec:4.2}
When predicting future states of a surrounding vehicle, not only the current state but also historical information needs to be considered~\cite{deo2018convolutional}. In this work, 
we use the acceleration prediction results from Section~\ref{sec:4.1} to dynamically update the state transition probability matrix at each time step.

 
 The system state $i$ of the surrounding vehicle is represented as a tuple with four discretized elements, including two-dimensional vehicle positions and velocities. The system input is expressed as a two-dimensional acceleration $(a_{1},a_{2})$. Note the current state probability is known in advance. Typically there is an initial state $i$ with $p_{i}(0)=1$, or an initial probability distribution is provided to address state uncertainties. In practice, from the current time, we need to calculate multiple stochastic FRSs at multiple forwarded time steps, and check the corresponding FRS at each future time step $k \in \{ 1,2,\dots,e\}$. 
 
 At each predicted time step $k$,  the acceleration prediction model provides a bivariate normal distribution  function $f_{m}^{k}(a_{1},a_{2})$ for each maneuver mode $m$ as

\begin{equation}
\begin{split}
&f^{k}_{m} (a_{1},a_{2})= \frac{1}{2\pi\sigma_{1}\sigma_{2}\sqrt{1-\rho^{2}}}\cdot \\ &\text{exp} \left( -\frac{1}{2(1-\rho^{2})} \left[\left ( \frac{a_{1}-\mu_{1}}{\sigma_{1}}\right)^2 + \left ( \frac{a_{2}-\mu_{2}}{\sigma_2}\right)^2 - 2\rho\frac{(a_{1}-\mu_{1})(a_{2}-\mu_{2})}{\sigma_{1}\sigma_{2}}\right ]  \right) 
\label{eq:PDF}
\end{split}
\end{equation}
where $\mu_{1},\mu_{2},\sigma_{1},\sigma_{2},\rho$ provided by the prediction model  denote predicted means and standard deviations along two directions, and the correlation at  future  time  instant $k$ for each maneuver mode $m$, respectively. The time and maneuver indices of the five parameters are omitted here for the sake of brevity.

 To propagate the system states, the conditional probability $p^{u}_{i}(k)$ at time step $k$ under state $i$ and acceleration $\mathbf{u} = (a_{1}^{u},a_{2}^{u})$ is calculated as 
 
 \begin{equation} \label{eq:u}
 {p}_{i}^{u}(k) = \frac{\widetilde{p}_{i}^{u}(k)}{\sum\limits_{u \in\mathcal{U}  } \widetilde{p}_{i}^{u}(k) }
 \end{equation}
 
 \begin{equation} \label{eq:u-assist}
     \widetilde{p}_{i}^{u}(k) =  \sum\limits_{m} \lambda_{m}^{k}\cdot \int_{\underline{a}_{2}^{u}}^{\overline{a}_{2}^{u}} \int_{\underline{a}_{1}^{u}}^{\overline{a}_{1}^{u}} f_{m}^{k}(a_{1},a_{2}) da_{1} da_{2}
 \end{equation}
 where $\lambda_{k}^{m}$ is the probability for maneuver mode $m$ at time step $k$, and $\underline{a}_{1}^{u},\overline{a}_{1}^{u},\underline{a}_{2}^{u},\overline{a}_{2}^{u}$ are the integral boundaries of $u$.
 
 Here the conditional state probability $p^{u}_{i}(k)$ is implicitly relevant to the current state as well as historical states. This is because the current and historical information has been considered when providing the predicted acceleration results. This implies the state transition  matrix now has to be computed online. 
 
 Substituting Eqs.~\eqref{eq:u} and~\eqref{eq:phi_u} into Eq.~\eqref{eq:phi}, the overall state transition matrix $\Phi$ is obtained. To distinguish the Markov-based approach which can compute the transition matrix offline, we denote the state transition matrix obtained with the prediction model at the predicted time step $k$ as $\Phi(k)$. Then at each predicted time step, the state probability vector is iteratively computed as 
 
  \begin{equation} \label{eq:transition-t}
    \mathbf{p}({k+1}) = \Phi(k)\cdot \mathbf{p}(k)
 \end{equation}

To measure the driving risk, the collision probability at the current time is expressed as the product of collision  probability at each predicted time step:
  \begin{equation}
        P_{col} = 1 -\prod \limits_{k} \left (1-\sum\limits_{i \in  \mathcal{H}(k)}p_{i}(k)\right)
 \end{equation}
 where $\mathcal{H}(k)$ is the set of states that  the ego vehicle position occupies at time step $k$. The vehicle dimension is considered when calculating the collision probability.

 \subsection{Infusing confidence belief}
 
 A prediction model applied in computing the stochastic FRS cannot always be accurate, since the vehicles could move unexpectedly. To address this issue, a confidence-aware belief vector is adopted to  modify probabilistic motion predictions that exploit modeled structure when the structure successfully explains vehicle motion, and degrade gracefully whenever the vehicle moves unexpectedly~\cite{fridovich2020confidence}.

If the prediction confidence is not considered, the probabilistic acceleration is expressed as in Eq.~\eqref{eq:PDF}.
To address the prediction confidence level, similar to~\cite{fridovich2020confidence}, an additional coefficient $\beta$ is infused into the acceleration probability density function:

\begin{equation}
\begin{split}
&f^{k}_{m} (a_{1},a_{2},\beta)= \frac{1}{2\pi(\sigma_{1}\beta)(\sigma_{2}\beta)\sqrt{1-\rho^{2}}}\cdot \\ &\text{exp} \left( -\frac{1}{2(1-\rho^{2})} \left[\left ( \frac{a_{1}-\mu_{1}}{\sigma_{1}\beta}\right)^2 + \left ( \frac{a_{2}-\mu_{2}}{\sigma_2\beta}\right)^2 - 2\rho\frac{(a_{1}-\mu_{1})(a_{2}-\mu_{2})}{\sigma_{1}\beta\sigma_{2}\beta}\right ]  \right) 
\label{eq:PDF-beta}
\end{split}
\end{equation}
where a positive coefficient $\beta$ controls the confidence level of the prediction model. 
For instance, when setting $\beta$ to infinity, the acceleration probability would be uniformly distributed and ignore the prediction model. While $\beta$ is close to 0, the discretized input $(a_{1},a_{2})$, which is \blue{closest} to the predicted acceleration mean values $(\mu_{1},\mu_{2})$, is assigned with probability one. 

Consequently, the input conditional probability computation in Eq.~\eqref{eq:u-assist} is extended as
 \begin{equation} \label{eq:u-betas-assist}
     \widetilde{p}^{u}(k,\beta) =  \sum\limits_{m} \sum\limits_{\beta}  \lambda_{m}^{k}\cdot b^{k}(\beta)\cdot \int_{\underline{a}_{2}^{u}}^{\overline{a}_{2}^{u}} \int_{\underline{a}_{1}^{u}}^{\overline{a}_{1}^{u}} f_{m}^{k}(a_{1},a_{2},\beta) da_{1} da_{2}
 \end{equation}
 where $b^{k}(\beta) \in (0,1)$ is the belief value for a specific $\beta$ at time step $k$.
 

Note that the performance of a prediction model possibly changes over time. For instance, the  model could have a relatively worse prediction accuracy when the surrounding vehicle starts to maneuver a lane change, as the lane-change maneuver may not be timely recognised. To reflect the dynamic property of the confidence level, the belief of each confidence level should be also updated frequently. In doing so, a Bayesian belief vector regarding possible values of $\beta$ is introduced. 

Initially, each $\beta$ is assigned with a uniform probability $b^{0}({\beta})$; then the belief vector evolves given posterior probabilities of the state and input under each $\beta$ at  time step $k$:

 \begin{equation}
    b^{k+1}(\beta) = \frac{P(\mathbf{u}^{(k-k^{\prime}+1:k)}|\mathbf{x}^{(k-k^{\prime}+1:k)}; \Theta^{(k-k^{\prime}+1:k)},\beta) b^{k}(\beta)}  {\sum\limits_{\widetilde{\beta}}P(\mathbf{u}^{(k-k^{\prime}+1:k)}|\mathbf{x}^{(k-k^{\prime}+1:k)}; \Theta^{(k-k^{\prime}+1:k)},\widetilde{\beta}) {b}^{k}(\widetilde{\beta})}
 \end{equation}
 where 
  \begin{equation}\label{eq:beta_update}
    P(\mathbf{u}^{({s}:k)}|\mathbf{x}^{({s}:k)}; \Theta^{({s}:k)},\beta) = \prod \limits_{i=k-k^{\prime}+1}^{k} \widetilde{p}^{u_i}({k-i},\beta)
 \end{equation}
 is the posterior probability of the observed actual accelerations $\mathbf{u}^{({s}:k)}$ with $t_{s} = t_{\text{max}(k-k^{\prime}+1,1)}$ from previous $k^{\prime}$ time steps. 
 
 
 In practice, the Bayesian belief vector with a relatively small set and small previous time steps, e.g., 5 discrete values of $\beta$ and $k^{\prime} = 2$, can achieve significant improvemente~\cite{fridovich2020confidence}.

\section{Integrated collision detection framework} \label{sec:frame}

In this section, we propose an integrated collision detection framework combing BRS and stochastic FRS described in the previous section. To this end, we also specify  the system dynamics models for the BRS and stochastic FRS respectively, and prove the equivalence of different models in  the BRS and stochastic FRS.

\subsection{Framework}

\begin{figure}[h!]
	\begin{center}		\includegraphics[width=1\textwidth]{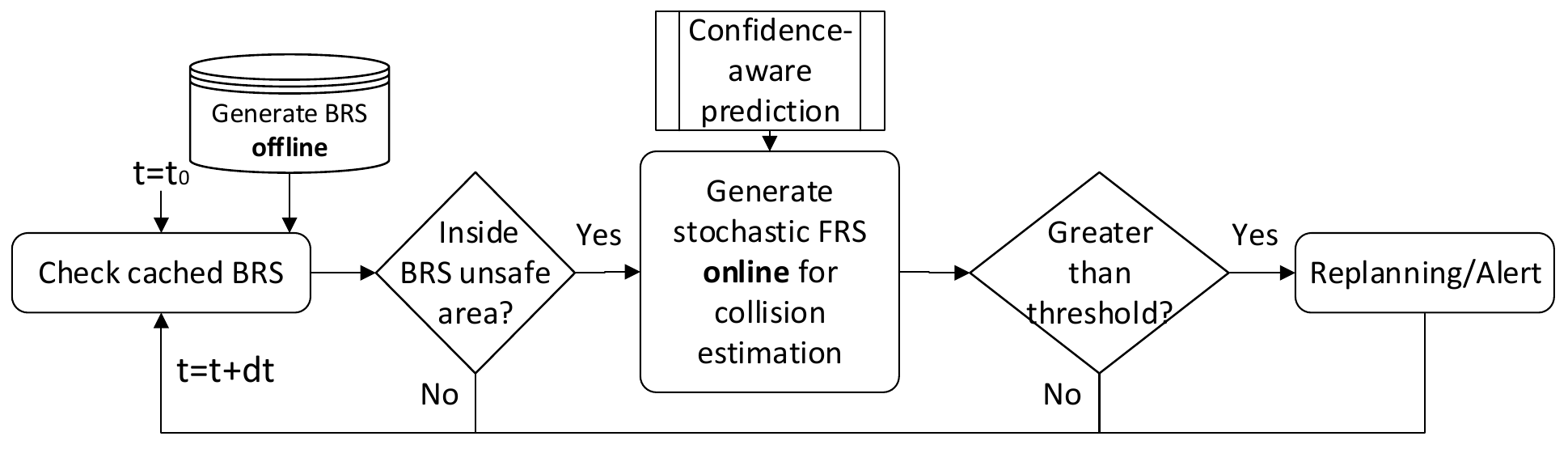}
		\caption{Flowchart of the integrated  collision detection framework.
}\label{fig:framework}
	\end{center}
\end{figure}

We propose an integrated  collision detection framework on highways by combining the BRS and stochastic FRS. 
Before employing the framework, a BRS is computed based on the HJI PDE~\cite{li2020prediction}. Although computing a BRS is time-consuming, this can be done offline and the results of BRS are cached in a look-up table for real-time risk assessment later. Note that we need to interpolate the BRS look-up table in practice, as the cached states are \blue{discretized}. The flowchart of the framework including three iterative steps is illustrated in Fig.~\ref{fig:framework}. 
  \begin{itemize}
  \item [Step 1] At the current time, we check whether the surrounding vehicle is inside the unsafe area identified by the BRS. If so, go to the next step; otherwise, the safety of vehicle interactions is theoretically ensured at the current time.
  \item [Step 2] A stochastic FRS is generated online to calculate collision probability at each predicted time step. Note that the stochastic FRS can be obtained using either heuristic rules~\cite{althoff2009model} and a prediction-based approach that has been introduced in Section~\ref{sec:models}.
  \item [Step 3] If the obtained collision probability is above a pre-defined threshold, the ego vehicle has to replan motion trajectories or the ego driver will receive an alert, to avoid potential crashes with the surrounding vehicle, which is not within the scope of this work. 
 \end{itemize}
 
\subsection{Vehicle dynamics and equivalent transformation between BRS and stochastic FRS}\label{sec:dynamics}

The computation of BRS and (stochastic) FRS depends on the selection of  vehicle dynamic models. Ideally, we can use the same system dynamics for the ego to construct both the BRS and FRS; then the equivalent transformation, which is used to ensure a consistent control input range between different vehicle dynamics, is no longer necessary. However, in this work, we use different system dynamics for the BRS and FRS due to two reasons. First, we construct the BRS using a bicycle model for the ego vehicle and a unicycle model for the surrounding vehicle following~\cite{li2020prediction}, since these models are relatively more realistic than a point mass model. Second, we construct the FRS using a point mass model, mainly to accommodate control input predictors, which typically output two-dimensional acceleration values~\cite{zhou2017recurrent,su2020graph}. Details of the employed vehicle dynamics in this work are as follows.

\begin{figure}[!htbp]
	\centering
	\subfigure[Ego vehicle dynamic model with inputs: acceleration  $a_{ego}$ and steering angle $\delta_{f}$.]{
		\includegraphics[width=0.47\textwidth]{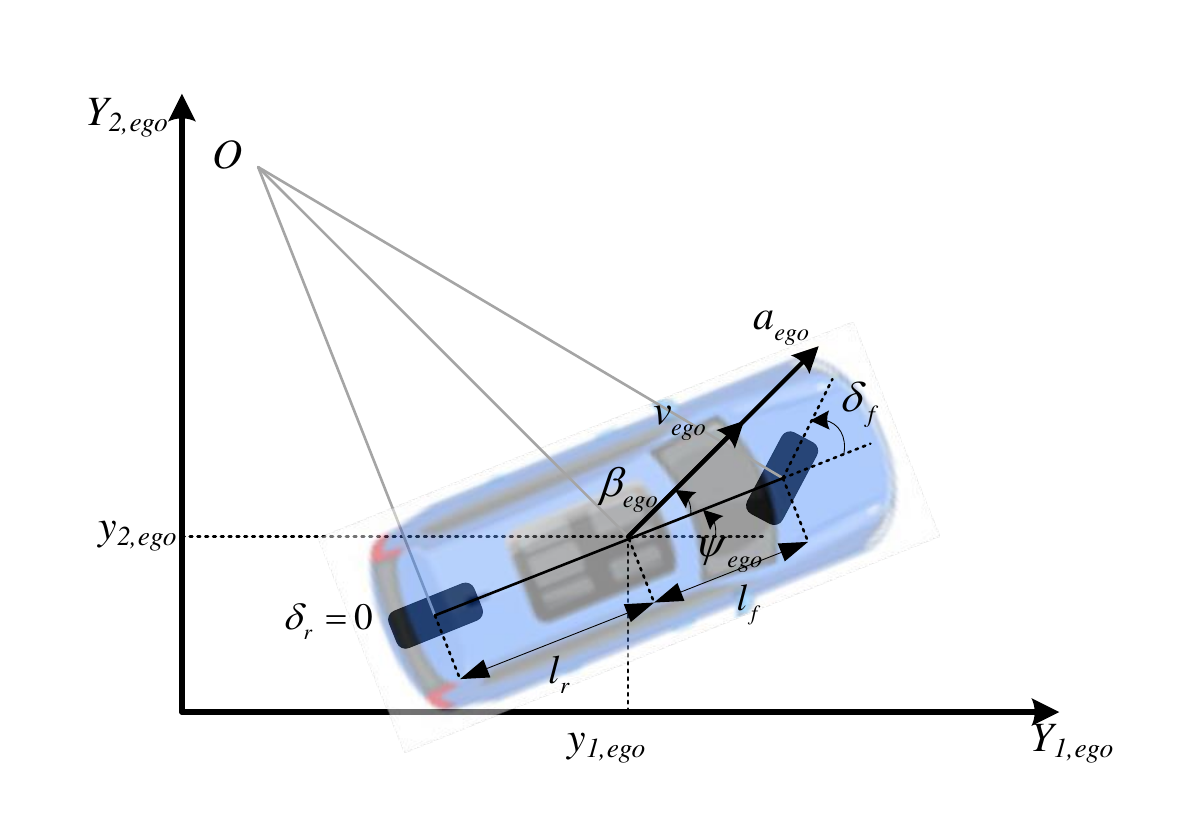}
	}
	\quad
		\subfigure[Surrounding vehicle dynamic model with inputs: acceleration $a_s$ and angular acceleration $\omega_{s}$.]{
		\includegraphics[width=0.47\textwidth]{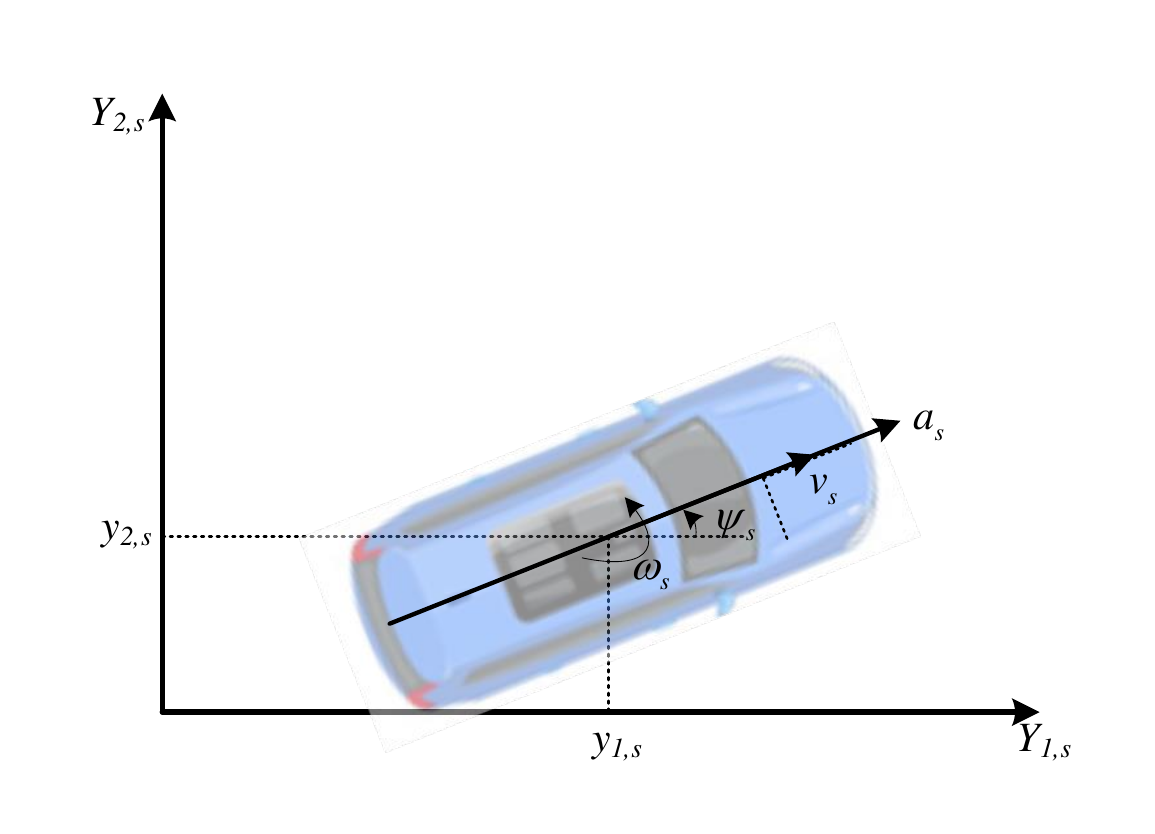}
	}
\caption{Vehicle dynamic systems in the BRS. }\label{fig:RS_dynamics}
\end{figure}

To calculate the BRS, the vehicle dynamics of ego and surrounding vehicles are respectively defined in line with~\cite{li2020prediction}. \blue{A more complex ego vehicle model will increase the dimensions of the relative dynamics model, which results in an overly complicated calculation of the BRS. In~\cite{leung2020infusing}, computing the BRS with a 7-dimensional relative system and discretization takes approximately 70 hours on a 3.0GHz octocore AMD Ryzen 1700 CPU. Meanwhile, in this work, collision detection is guaranteed by the combination of the BRS and the stochastic FRS. The vehicle models used by the two modules (BRS and stochastic FRS) need to be consistent or similar. A high-dimensional vehicle model in the BRS will significantly increase the computational load of stochastic FRS, even if with a parallel computation assumption. Therefore,} as shown in Fig.~\ref{fig:RS_dynamics}, we apply a  bicycle model for the ego vehicle, where the state includes longitudinal/lateral positions $y_{1,ego}/y_{2,ego}$, the heading angle $\psi_{ego}$, and the ego vehicle velocity ${v}_{ego}$. Its control input is the acceleration $a_{ego}$ and steering angle $\delta_{f}$. $O$ is the vehicle rotation center. $\delta_{r}=0$ is the rear wheel steering angle and $\beta_{ego} = \text{tan}^{-1}(\frac{l_r}{l_{f}+l_{r}}\text{tan}(\delta_{f}))$ is the slip angle of the ego vehicle, where $l_{f}/l_{r}$ is the distance from front/rear to the vehicle reference point. For the surrounding vehicle, we model its state with two-dimensional positions $y_{1,s}/y_{2,s}$, the heading angle $\psi_{s}$, and the surrounding vehicle velocity ${v}_{s}$. The control input is acceleration $a_s$ and angular velocity $\omega_{s}$. Then the relative dynamics can be represented as

\begin{equation}
\left\{
\begin{aligned}
   & \dot{y}_{1,rel} = \frac{v_{ego}}{l_r}\text{sin}\beta_{{ego}}\cdot y_{2,rel} + v_{s} \text{cos} \psi_{rel} - v_{{ego}} \text{cos}\beta_{ego} \\
      &  \dot{y}_{2,rel} = -\frac{v_{ego}}{l_r}\text{sin}\beta_{{ego}}\cdot y_{1,rel} + v_{s}\text{sin} \psi_{rel} - v_{{ego}} \text{sin}\beta_{{ego}} \\
       & \dot{\psi}_{rel} = \omega_{s} - \frac{v_{{ego}}}{l_{r}}\text{sin}\beta_{{ego}}\\
       & \dot{v}_{ego} = a_{{ego}}\\
       & \dot{v}_s = a_{s}
        \end{aligned}
        \right.
\end{equation}
where $y_{1,rel}/y_{2,rel}$ and $\psi_{rel}$ are relative two-dimensional coordinates and heading angel respectively.  

For the calculation of  the stochastic FRS, to accommodate acceleration prediction models that typically represent probabilistic accelerations with bivariate distributions, we employ a point mass model to compute the stochastic FRS of the surrounding vehicle. The control input  is simplified with two-dimensional accelerations, 
and the future vehicle positions can then be directly propagated with the predicted accelerations. It is assumed that the planned trajectories of the ego vehicle are deterministic and known in advance. Nevertheless, the motion uncertainties of the ego vehicle can be represented by extending its future position with additional adjacent states, which is left for future work. 

\begin{table}[htbp]
\scriptsize
  \centering
  \caption{System dynamics to construct the BRS and stochastic FRS.}
    \begin{tabular}{l p{16em} p{15em}}
    \toprule
          & \multicolumn{1}{c}{BRS} & \multicolumn{1}{c}{FRS} \\
    \midrule
    ego   & Bicycle model with input:\newline{}acceleration and steering angle  & / \\
    sur   & Unicycle model with input:\newline{}acceleration  and angular acceleration  & {Point mass model with input:\newline{}two-dimensional accelerations} \\
    \bottomrule
    \end{tabular}%
  \label{tab:system_dynamic}%
\end{table}%

A summary of the system dynamics used for the BRS and FRS is provided in Table~\ref{tab:system_dynamic}. Given different  vehicle dynamic models and control inputs employed for the BRS and stochastic FRS, it is essential to match the control input range among the three dynamic models. We call this process the equivalent vehicle dynamics transformation between the BRS and FRS. We provide a detailed equivalent transformation procedure in the Appendix. Now the BRS and stochastic FRS can be integrated into the same framework.

\section{Experiments} \label{sec:sim}
In this section, we first introduce the employed naturalistic highway driving dataset highD for the prediction model training/testing, as well as the experimental setup for the prediction model and the BRS/FRS computation. Then a number of experiments are designed and conducted with respect to three aspects: 1) We test the proposed acceleration prediction  model by comparing it with three existing predictors, since the acceleration prediction model plays a vital role in the establishment of the stochastic FRS. 2) We further test the performance of the prediction-based confidence-aware FRS, using both naturalistic driving data from highD, as well as simulated cut-in events. The risky cut-in events are simulated to test collision estimation performance based on the stochastic FRS. 3) In order to validate the integrated collision detection framework, we first compare the identified unsafe areas between the BRS and FRS, and then test the framework in both risky and  non-risky cut-in events. 


\subsection{Dataset and setup}\label{sec:setup}

The highD dataset~\cite{krajewski2018highd}, which contains bird-view naturalistic driving data on German highways, is utilized to train and test the acceleration prediction model. We randomly select equal samples for the three different lane-change maneuver modes, leading to 
135,531 (45,177 for each maneuver mode) and 19,482 (6,494 for each mode) samples for the training and testing respectively. The original dataset sampling rate is 25 Hz, and we downsample by a factor of 5 to reduce the model complexity. We consider 2-seconds historic information as input and predict within a 2-second horizon.  

The prediction model is trained using Adam with a learning rate 0.001, and the sizes of the encoder and decoder are 64 and 128 respectively. The size of the fully connected layer is 32. The convolutional social pooling layers consist of a $3\times 3$ convolutional layer with 64 filters, a $3\times 1$ convolutional layer with 16 filters, and a $2\times 1$ max pooling layer, which are consistent with the settings in~\cite{deo2018convolutional}. 

In line with the vehicle dynamics in Section~\ref{sec:dynamics}, the BRS is offline computed in five dimensions, among which are relative two-dimensional positions and heading angle, and vehicle velocities for the ego and the surrounding, respectively. \blue{The relative longitudinal position is discretized from -10 to 40 meters with an increment 0.5  meters, the relative lateral position from -4 to 4 meters with an increment 0.4 meters,}  the heading angle from -45 to 45 degrees with an increment 9 degrees, and the ego/surrounding vehicle velocity from 20 to 40 m/s with an increment 1 m/s, leading to over five million states. \blue{The longitudinal position range setting is sufficient to identify an enclosed unsafe area for all simulations in this work, as shown as an example in Fig.~\ref{fig:BRSvsFRS} later.} The range of control input is set in the Appendix to ensure equivalent model transformation between the BRS and FRS.

The FRS states are expressed in four dimensions, including two dimensions for position and velocities, respectively. \blue{The vehicle longitudinal position is discretized from -4 to 80 meters with an increment 2 meters, and the relative lateral position is from -4 to 4 meters with an increment 1 meter. The longitudinal velocity is discretized from 20 to 40  m/s with an increment 0.4  m/s, and the lateral velocity is from -2.5 to 2.5 m/s with an increment 0.2 m/s,} leading to around half a million states. \blue{Here the longitudinal position range of FRS is from -4 to 80 meters, because we assume a vehicle cannot move backwards, and the marginal position to collide with a surrounding vehicle behind is -4 meters given the vehicle length (4 meters). Meanwhile, it can reach a maximum of 80 meters with the largest longitudinal velocity (40 m/s) in the prediction horizon (2 seconds).}  As for the control input, we discretize the longitudinal (lateral) accelerations  from -5 to 3 (-1.5 to 1.5) m/s$^2$ with an increment 1 (0.5) m/s$^2$, leading to 63 acceleration combinations. We also add several constraints to limit the acceleration selection, including maximal acceleration, strict forward motion, and maximal steering angle~\cite{mullakkal2020probabilistic}. In the end, 37 million possible state transfers are generated. To alleviate the computational load, we assume that an advanced GPU~\cite{turner2018application}, which enables 2048$\times$28 parallel 
computation, is available. The stochastic FRS with state probability distributions $\mathbf{p}({{k}})$ is calculated at each predicted future time step within 2 seconds with an increment 0.4 seconds, i.e., $\{0.4, 0.8, 1.2, 1.6, 2.0\}$.

\subsection{Acceleration/trajectory prediction models}

To validate the proposed two-stage acceleration prediction  model (denoted as T-LSTMa), we compare T-LSTMa with three \blue{state-of-the-art} probabilistic multi-modal predictors: the two-stage trajectory prediction model T-LSTM~\cite{wang2021probabilistic}, the social convolutional trajectory predictor S-LSTM using convolutional neural networks to represent surrounding vehicles~\cite{deo2018convolutional}, and its variation S-LSTMa \blue{for fair comparisons, where the prediction output is modified to probabilistic accelerations.}
Utilizing the testing dataset from highD, we report the comparative results in Table~\ref{tab:tracks}, including five evaluation indicators, i.e., root mean square error (RMSE), average displacement error (ADE), final displacement error (FDE), NLL (the lower, the better), and average lane-change prediction F1 score \blue{LC-F1 (a metric calculated as the harmonic mean of the precision and recall~\cite{goutte2005probabilistic})}. We show predictor performance on the overall testing dataset, and also compare prediction results in terms of three lane-change maneuver modes, respectively. Note that we do not directly evaluate the acceleration prediction accuracy. Instead, we compare vehicle position prediction accuracy, which directly affects the collision estimation.


\begin{table}[htbp]
\scriptsize
  \centering
  \caption{Prediction model performance on highD testing dataset.}
    \begin{tabular}{ccrrrr}
    \toprule
          &  Indicators     & \multicolumn{1}{c}{T-LSTMa} & \multicolumn{1}{c}{S-LSTMa} & \multicolumn{1}{c}{T-LSTM} & \multicolumn{1}{c}{S-LSTM} \\
    \midrule
    \multicolumn{1}{c}{\multirow{5}[2]{*}{Lane-\newline{}Keeping}} & RMSE (m) & 0.17  & 0.42  & 0.46  & 0.59  \\
          & ADE  (m) & 0.10  & 0.31  & 0.26  & 0.27  \\
          & FDE  (m) & 0.26  & 0.81  & 0.61  & 0.72  \\
          & NLL  (m) & -3.47  & 0.91  & -1.59  & -1.38  \\
          & LC-F1 (\%) & 98.70  & 0.00  & 98.70  & 96.94  \\
    \midrule
    \multicolumn{1}{c}{\multirow{5}[2]{*}{Turning- \newline{}Left}} & RMSE (m) & 0.24  & 0.56  & 0.52  & 0.62  \\
          & ADE  (m) & 0.14  & 0.38  & 0.35  & 0.36  \\
          & FDE  (m) & 0.37  & 0.96  & 0.83  & 0.89  \\
          & NLL  (m) & -2.80  & -0.33  & -0.63  & -0.66  \\
          & LC-F1 (\%) & 99.64  & 87.04  & 99.64  & 97.85  \\
    \midrule
    \multicolumn{1}{c}{\multirow{5}[2]{*}{Turning- \newline{}Right}} & RMSE (m) & 0.23  & 0.29  & 0.49  & 0.58  \\
          & ADE  (m) & 0.14  & 0.17  & 0.33  & 0.37  \\
          & FDE  (m) & 0.37  & 0.45  & 0.74  & 0.86  \\
          & NLL  (m) & -2.97  & -2.25  & -0.74  & -0.46  \\
          & LC-F1 (\%) & 99.80  & 99.95  & 99.80  & 99.34  \\
    \midrule
    \multirow{5}[2]{*}{Overall} & RMSE (m) & \bf{0.22}  & 0.44  & 0.49  & 0.60  \\
          & ADE  (m) & \bf{0.13}  & 0.29  & 0.32  & 0.33  \\
          & FDE  (m) & \bf{0.34}  & 0.74  & 0.73  & 0.86  \\
          & NLL  (m) & \bf{-3.07}  & -0.58  & -0.97  & -0.82  \\
          & LC-F1 (\%) & \bf{98.79}  & 59.20  & \bf{98.79}  & 96.18  \\
    \bottomrule
    \end{tabular}%
  \label{tab:tracks}%
\end{table}%

Looking at the overall comparison results, we observe that using acceleration prediction and then propagating future vehicle position can significantly improve the prediction accuracy (see comparisons between S-/T-LSTMa and S-/T-LSTM). This could be due to the additional physical information when using the predicted acceleration to propagate future positions. Meanwhile, S-/T-LSTMa and S-/T-LSTM have the same LC-F1, as they share the same lane-change prediction submodel. S-LSTMa has the worst performance in terms of NLL and LC-F1. Specifically, the LC-F1 for lane-keeping trajectories is 0\%. This indicates jointly predicting acceleration and the lane-change maneuver mode in one neural network leads to undesirable results; it is reasonable to consider decoupling the acceleration and  lane-change maneuver mode predictions by two neural networks, which have been employed in T-LSTMa. To summarize, our proposed acceleration predictor T-LSTMa achieves the best performance in terms of all indicators, with ADE $<$ 0.15 meters and FDE $<$ 0.40 meters.

We further test the model prediction performance over different prediction horizons (one and three seconds). As shown in Table~\ref{tab:prediction_time}, the proposed predictor T-LSTMa 
achieves  superior performance over all prediction horizons. In addition, both models obtain more accurate prediction results with a shorter prediction horizon. For the remainder of this work, we use T-LSTMa with default settings to predict  accelerations.

\begin{table}[htbp]
\scriptsize
  \centering
    \caption{Prediction model performance over prediction horizon 1 and 3 seconds.}
\begin{tabular}{ccrrrr}\toprule
\multicolumn{1}{l}{}       & Indicators & \multicolumn{1}{c}{T-LSTMa} & \multicolumn{1}{c}{S-LSTMa} & \multicolumn{1}{c}{T-LSTM} & \multicolumn{1}{c}{S-LSTM} \\ \midrule
\multirow{5}{*}{1 second}  & RMSE (m)   & 0.06                        & 0.13                        & 0.15                       & 0.19                       \\
                           & ADE  (m)   & 0.04                        & 0.09                        & 0.14                       & 0.15                       \\
                           & FDE  (m)   & 0.08                        & 0.20                        & 0.25                       & 0.28                       \\
                           & NLL  (m)   & -4.79                       & -2.46                       & -2.31                      & -2.17                      \\
                           & LC-F1 (\%) & 98.90                       & 65.98                       & 98.90                      & 97.86                      \\\midrule
\multirow{5}{*}{3 seconds} & RMSE (m)   & 0.47                        & 0.86                        & 0.79                       & 0.85                       \\
                           & ADE  (m)   & 0.26                        & 0.56                        & 0.48                       & 0.53                       \\
                           & FDE  (m)   & 0.77                        & 1.52                        & 1.32                       & 1.48                       \\
                           & NLL  (m)   & -1.90                       & 1.11                        & -0.26                      & -0.02                      \\
                           & LC-F1 (\%) & 94.71                       & 52.18                       & 94.71                      & 91.00 \\\bottomrule               \label{tab:prediction_time}  \end{tabular}
\end{table}

\subsection{Confidence-aware position prediction and collision estimation}

 
In this section, we compare different approaches for generating stochastic FRS. Three different groups of coefficients $\boldsymbol{\beta}$ are selected: Only one distribution with $\boldsymbol\beta = [1]$; three normal distributions with $\boldsymbol\beta = [1/2, 1, 2]$; five normal distributions with $\boldsymbol\beta = [1/3, 1/2, 1, 2, 3]$. This leads to three different approaches to generating prediction-based stochastic FRS (denoted as PSRS, PSRS-3$\beta$ and PSRS-5$\beta$). Further increasing the number of distribution groups is not considered, since we need to ensure real-time computation of the \blue{stochastic FRS}. The heuristic method in~\cite{althoff2009model} (denoted as HSRS) is also adopted as a baseline to generate  the stochastic FRS.

To test the performance of the prediction-based confidence-aware FRS, we use both naturalistic driving data from highD, as well as simulated cut-in events. This is because highD itself does not contain safety-critical events, and we thus create simulated risky events to test collision estimation performance based on the stochastic FRS.

\subsubsection{highD trajectories}

Assume that the vehicle would occupy the exact space state and four adjacent states along with the longitudinal and lateral directions. We randomly select naturalistic driving trajectories from highD, including both  lane-keeping and the lane-change trajectories.


\begin{figure}[htbp]
	\centering
	\subfigure[HSRS]{
		\includegraphics[width=0.7\textwidth]{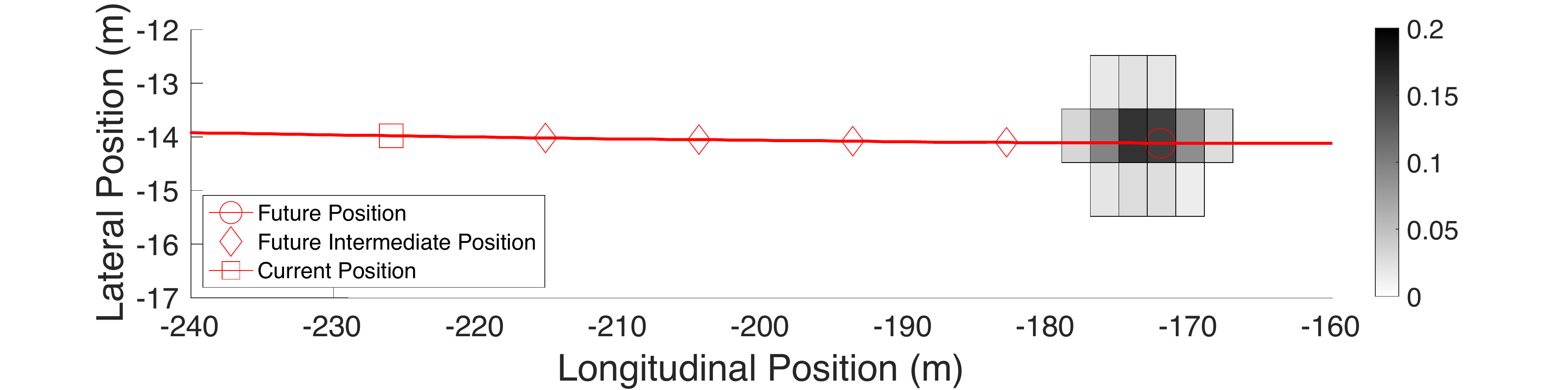}
	}
	\quad
		\subfigure[PSRS]{
		\includegraphics[width=0.7\textwidth]{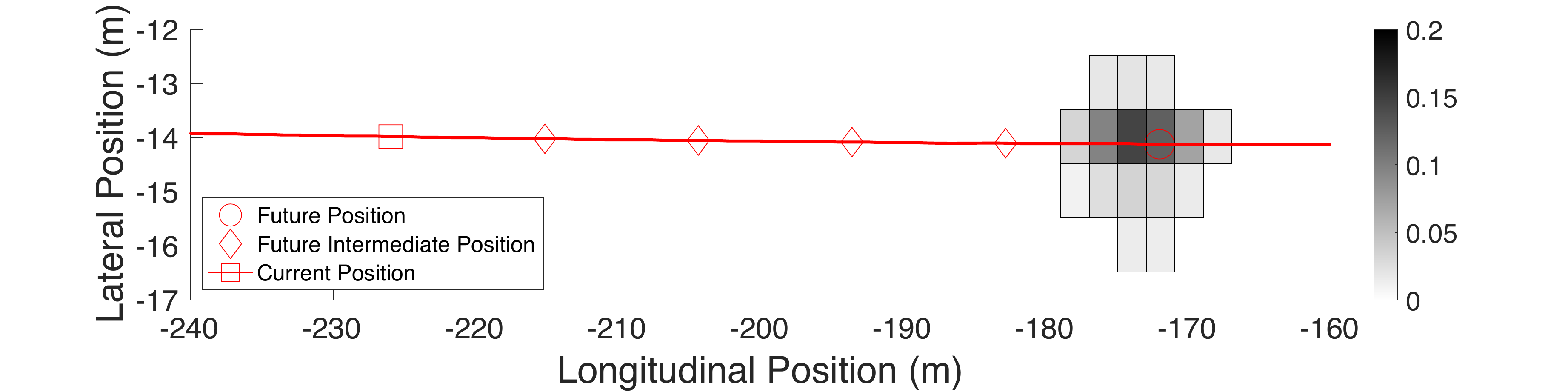}
	}
		\quad
	\subfigure[PSRS-3$\beta$]{
		\includegraphics[width=0.7\textwidth]{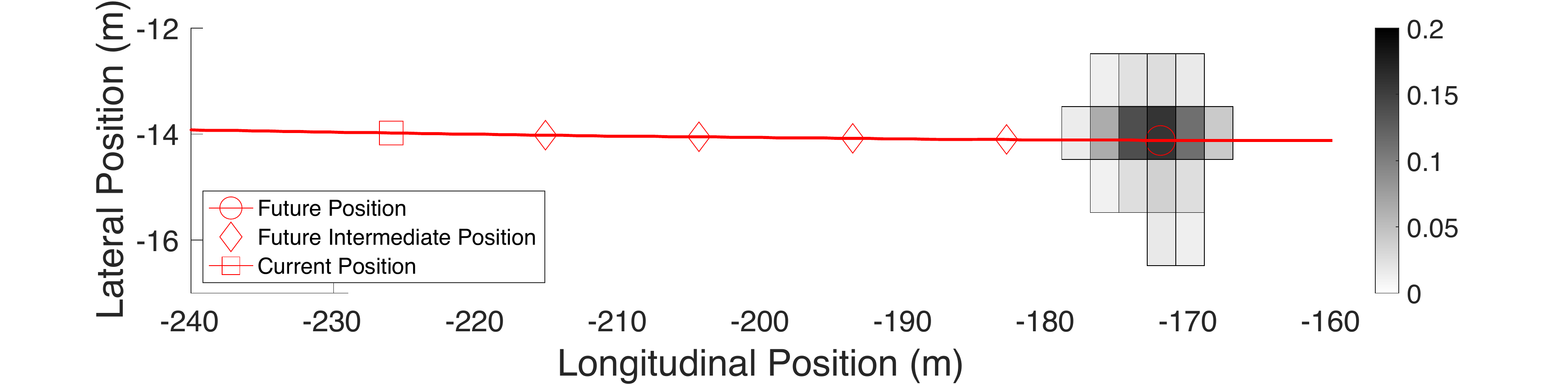}
	}
		\quad
	\subfigure[PSRS-5$\beta$]{
		\includegraphics[width=0.7\textwidth]{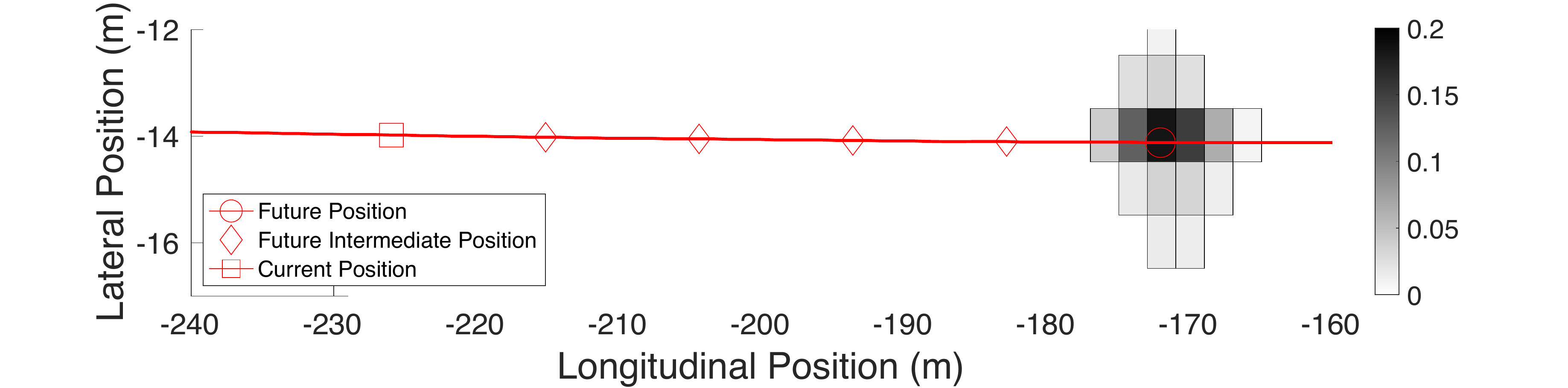}
	}
\caption{State probability distributions of the stochastic FRS using different approaches for a lane-keeping trajectory (the current time $t=4.4$ seconds). Only position states with $>$ 0.01 probability are displayed.}\label{fig:SFRS_LK}
\end{figure}

\begin{figure}[htbp]
	\centering
	\subfigure[HSRS]{
		\includegraphics[width=0.7\textwidth]{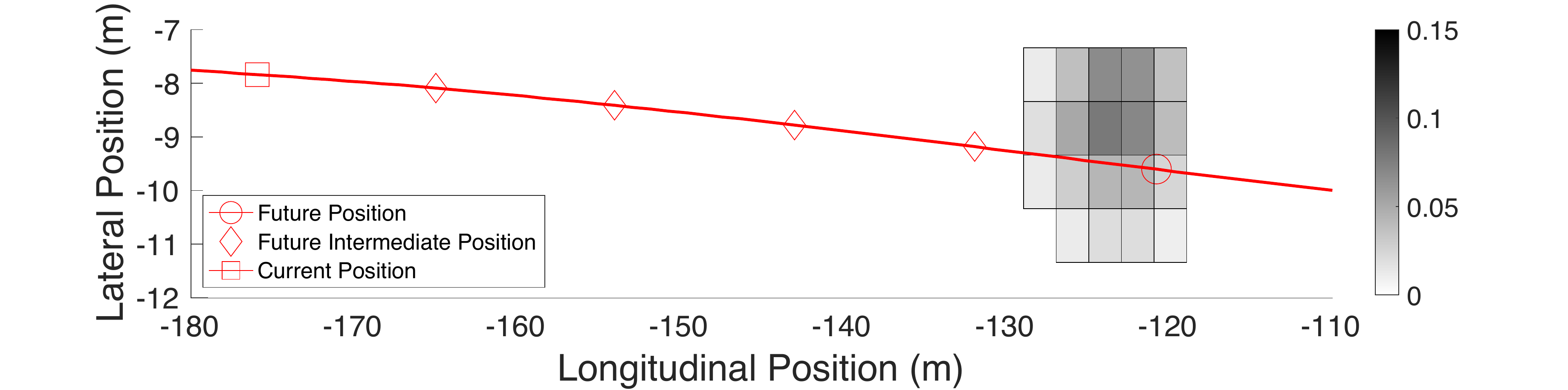}
	}
	\quad
		\subfigure[PSRS]{
		\includegraphics[width=0.7\textwidth]{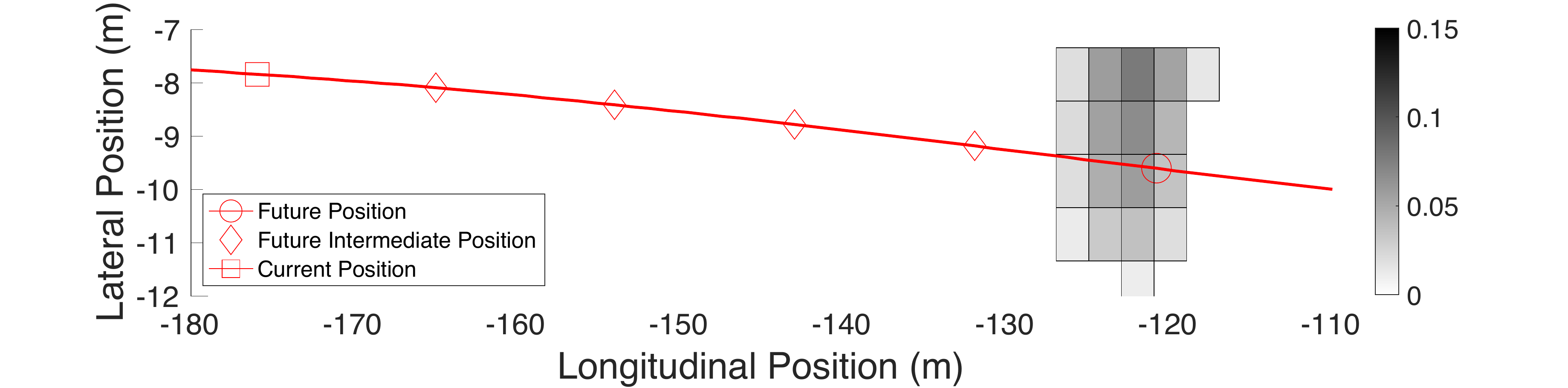}
	}
		\quad
	\subfigure[PSRS-3$\beta$]{
		\includegraphics[width=0.7\textwidth]{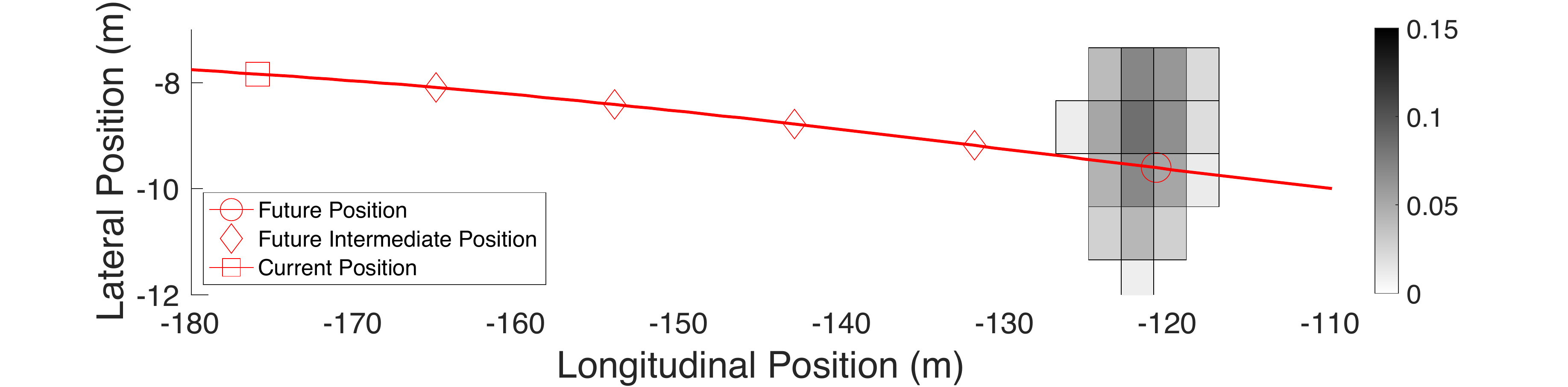}
	}
		\quad
	\subfigure[PSRS-5$\beta$]{
		\includegraphics[width=0.7\textwidth]{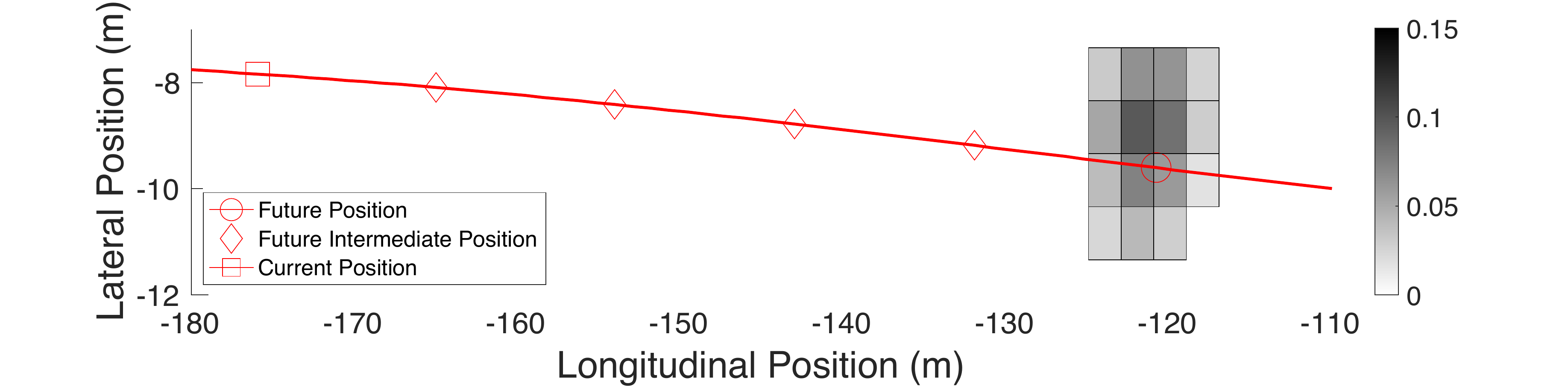}
	}
\caption{State probability distributions of stochastic FRS using different approaches during a lane change (the current time $t=6.4$ seconds). Only position states with $>$ 0.01 probability are displayed.}\label{fig:SFRS_LC}
\end{figure}

\begin{figure}[htbp]
	\centering
	\subfigure[Lane-keeping (PSRS-3$\beta$)]{
		\includegraphics[width=0.45\textwidth]{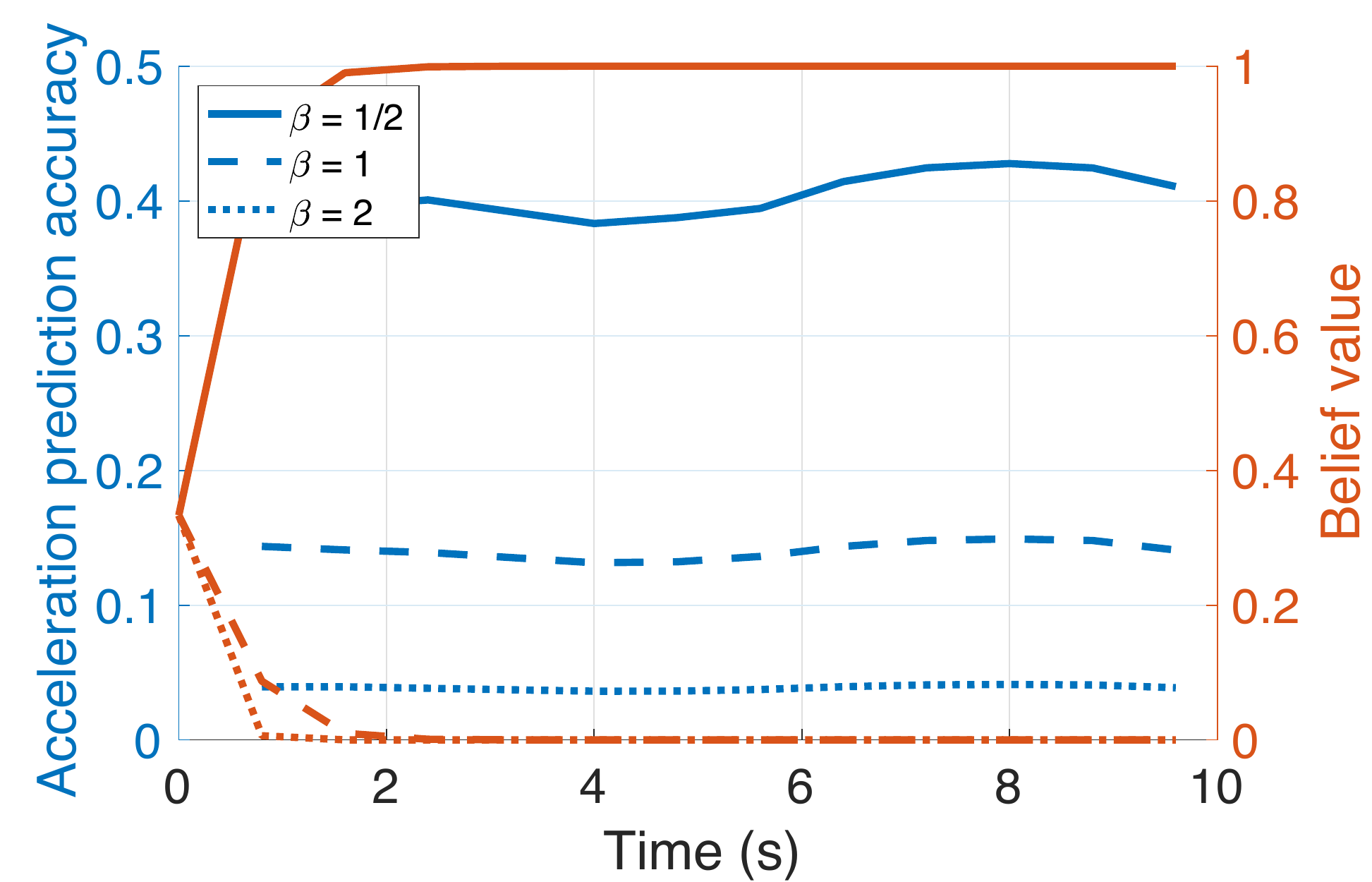}
	}
	\quad
		\subfigure[Lane-keeping (PSRS-5$\beta$)]{
		\includegraphics[width=0.45\textwidth]{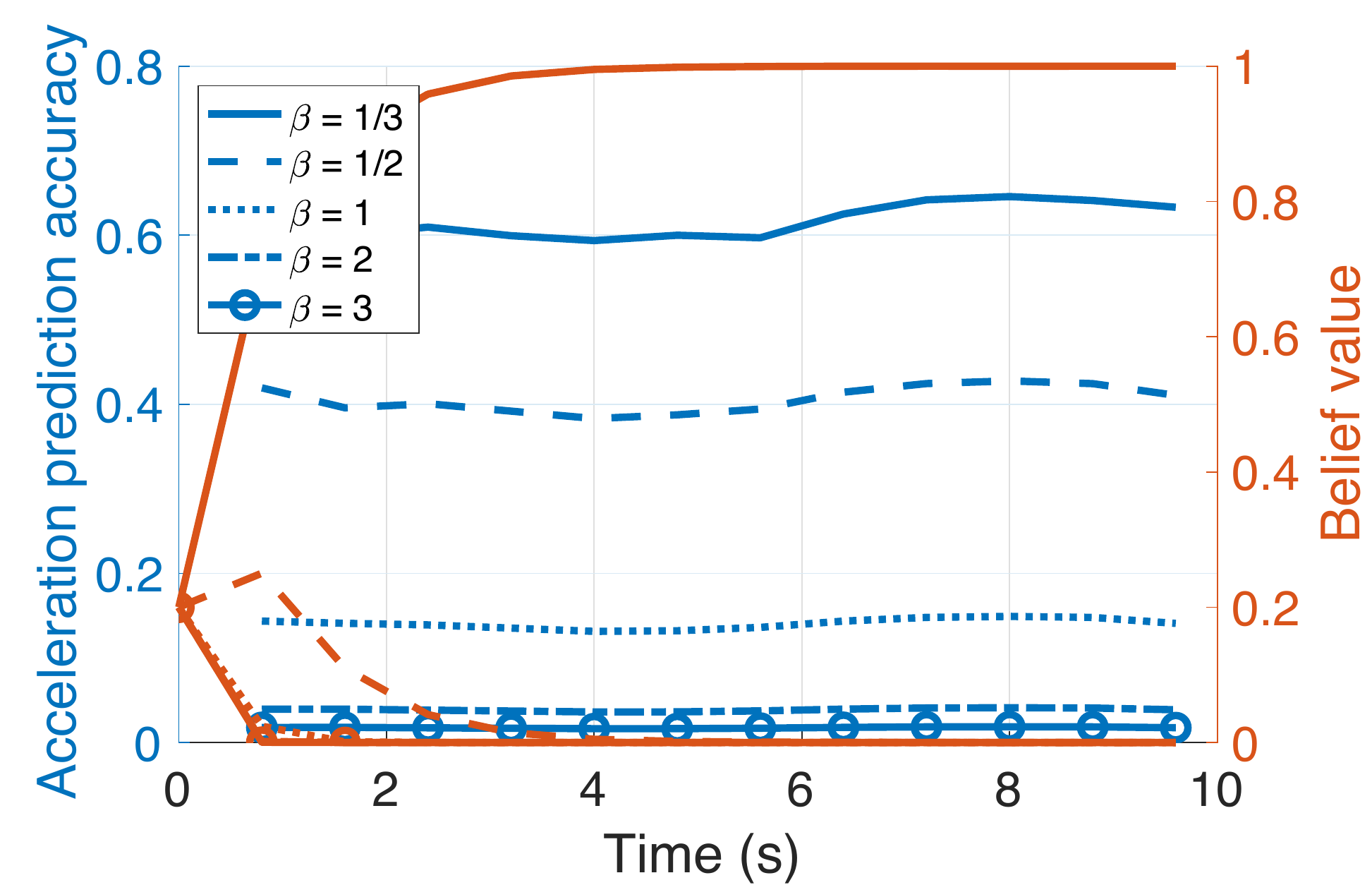}
	}
 \quad
		\subfigure[Lane-change (PSRS-3$\beta$)]{
		\includegraphics[width=0.45\textwidth]{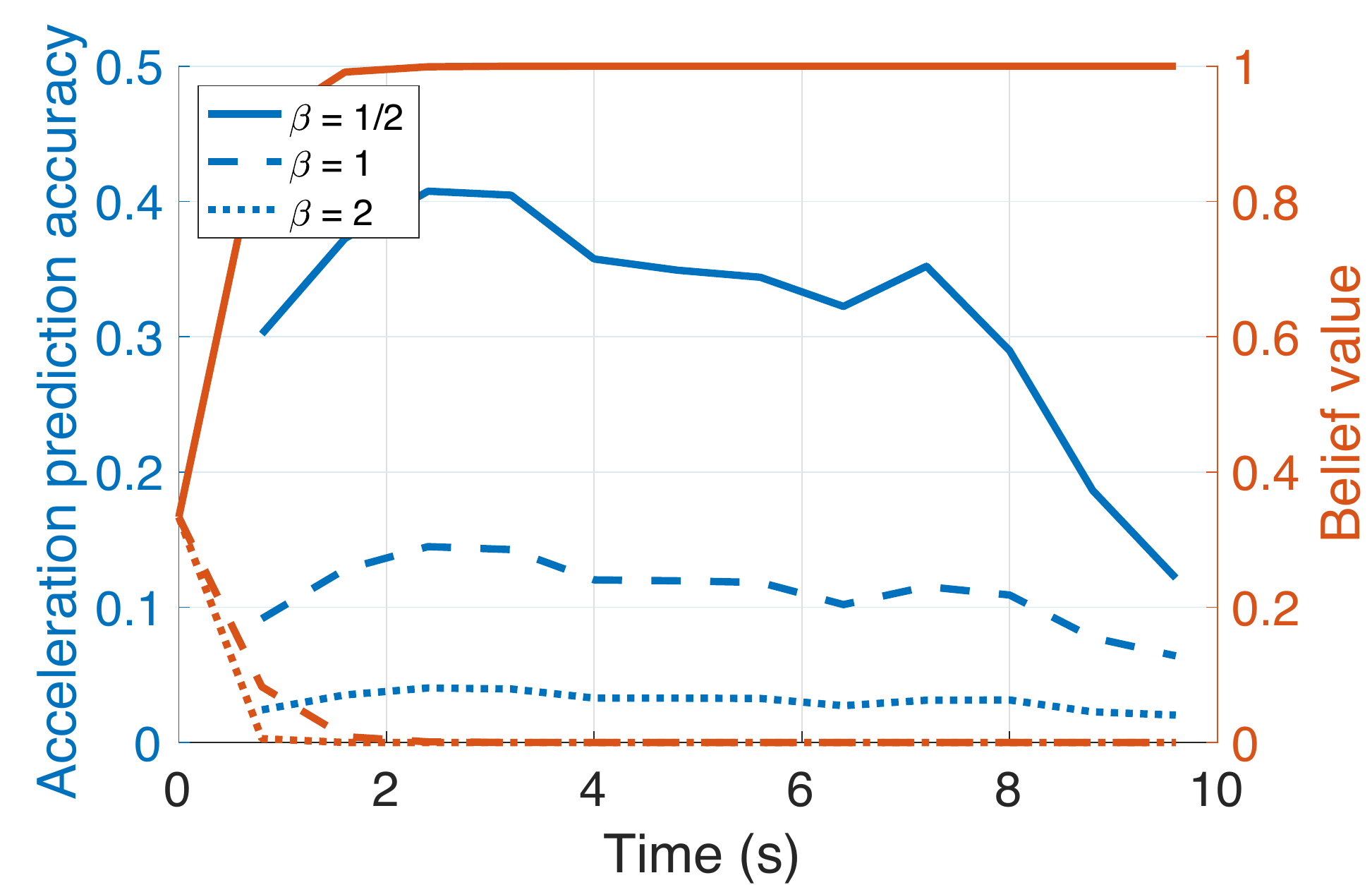}
	}
 \quad
		\subfigure[Lane-change (PSRS-5$\beta$)]{
		\includegraphics[width=0.45\textwidth]{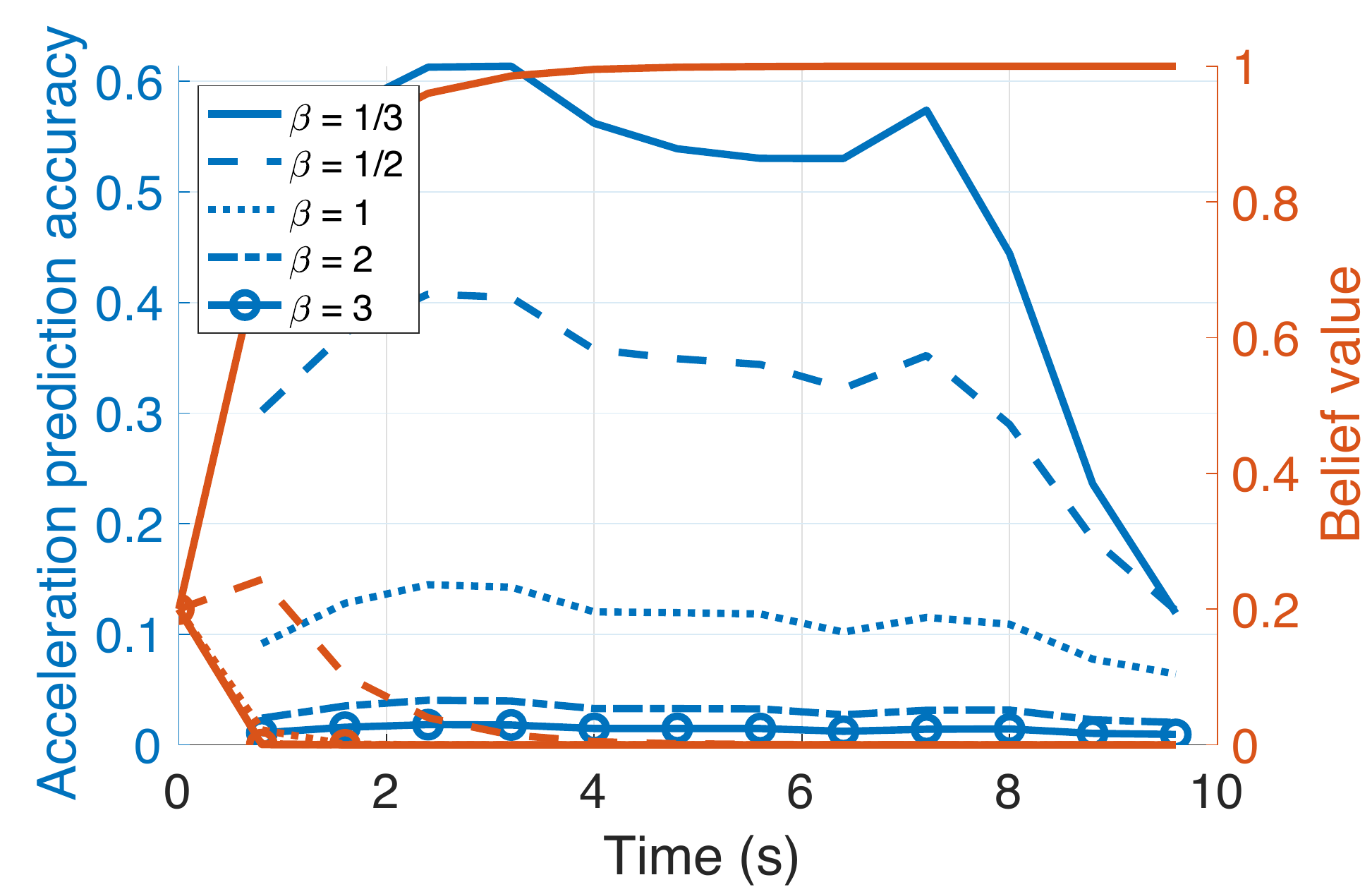}
	}
\caption{Belief dynamic changes \blue{using naturalistic driving data} under different coefficient settings. \blue{Subfigures (a) and (b) are for a lane-keeping trajectory, while subfigures (c) and (d) are for a lane-change trajectory. Blue lines represent acceleration prediction accuracy change with different coefficient settings, while orange lines indicate corresponding dynamic belief values.}}\label{fig:belief_update}
\end{figure}

Examples of predicted stochastic FRSs using different approaches are presented in Fig.~\ref{fig:SFRS_LK} and Fig.~\ref{fig:SFRS_LC} respectively. For the lane-keeping trajectory, the position prediction accuracy at time $t=4.4$ seconds of the four approaches is 44.33\%, 45.56\%, 46.84\%, and 52.74\%. For the lane-change trajectory, the position prediction accuracy at time $t=6.4$ seconds of the four approaches is 24.54\%, 29.38\%, 32.58\%, and 37.15\%. Infusing confidence awareness can indeed improve future vehicle position prediction, in particular for lane-change trajectories. 
The belief vector changes are illustrated in Fig.~\ref{fig:belief_update} for \blue{both lane-keeping and lane-change trajectories. For the lane-keeping trajectory, subfigures (a) and (b) corresponding results from PSRS-3$\beta$ and PSRS-5$\beta$ both have stable acceleration prediction accuracy, as the surrounding vehicle moves as expected. Besides, the belief value with the lowest $\beta$ converges to one for both two approaches. This is because the confidence-infused prediction model provides a more accurate mean value of accelerations, leading to a corresponding higher belief value with lower $\beta$ in line with Eq.~\eqref{eq:beta_update}. For the lane-change trajectory, a similar convergence trend of belief value changes can be observed in subfigures (c) and (d). However, the belief value of the lowest $\beta$ in subfigure (d) (PSRS-5$\beta$) takes about 4 seconds to converge while it takes less than 2 seconds in subfigure (c). This indicates that  infusing a higher-dimension belief vector is capable of adjusting more complicated prediction scenarios (e.g., a lane-change trajectory) and providing more accurate prediction results.}

We test the four approaches using  randomly selected 100 trajectories from highD for both lane-keeping and lane-change situations. The average position prediction performance is summarized in Table~\ref{tab:pos_pred}. Clearly when the prediction horizon is short (e.g.\@ 0.4 seconds), all approaches can generate the stochastic FRS with higher prediction accuracy. However, when the prediction horizon becomes larger, HSRS has the worst performance followed by PSRS. Again, infusing confidence belief effectively improved  the prediction accuracy; such improvement is more significant during lane-change situations.

\begin{table}[htbp]
\scriptsize
  \centering
  \caption{Average position prediction accuracy (\%)    using four stochastic FRSs with different prediction times. LK and LC stand for vehicle lane-keeping and lane-change situations respectively.}
    \begin{tabular}{crrrrrrrr}
    \toprule
          & \multicolumn{2}{c}{2.0 s} &       & \multicolumn{2}{c}{1.2 s} &       & \multicolumn{2}{c}{0.4 s} \\
\cmidrule{2-3}\cmidrule{5-6}\cmidrule{8-9}          & \multicolumn{1}{c}{LK} & \multicolumn{1}{c}{LC} &       & \multicolumn{1}{c}{LK} & \multicolumn{1}{c}{LC} &       & \multicolumn{1}{c}{LK} & \multicolumn{1}{c}{LC} \\
    \midrule
    HSRS  & 43.08  & 26.81  &       & 79.45  & 50.01  &       & 100.00  & 87.08  \\
    PSRS  & 45.44  & 30.47  &       & 78.38  & 50.48  &       & 100.00  & 86.22  \\
    PSRS-3$\beta$ & 48.64  & 34.90  &       & 80.23  & 51.69  &       & 100.00  & 86.95  \\
    PSRS-5$\beta$ & 52.81  & 38.58  &       & 84.01  & 55.63  &       & 100.00  & 88.39  \\
    \bottomrule
    \end{tabular}%
  \label{tab:pos_pred}%
\end{table}%

\subsubsection{Simulated safey-critical cut-in trajectories} \label{sec:sim_tracks}

Safety-critical cut-in events are simulated to test the collision detection performance on different stochastic FRSs. In the simulated cut-in scenario, the ego vehicle is in the middle lane and the surrounding travels in the right lane with constant initial longitudinal speeds $v_{e}=30$ m/s and $v_{s}=25$ m/s, respectively. The surrounding vehicle is 15 meters ahead of the ego at $t = 1.0$ second, and starts turning left with a lane-change duration 7.5 seconds. The vehicle length and width are 4 and 2 meters, respectively. The driving behaviors of the two vehicles are simulated by the classic intelligent driving model (IDM) for car following, and a lateral control model in~\cite{mullakkal2020hybrid}.

The IDM  equations are expressed as 
\begin{equation}\label{eq:IDM}
\left \{
\begin{aligned}{}
    s^{\star}(t) &= s_{0} + \text{max}(0,v(t) T + v(t) dv(t)/(2 \sqrt{a_{a} a_{b}}))  \\
a_{1}(t) &= \text{max}(a_{a} (1 - (v(t)/v_{0})^4 - (s^{\star}(t)/s(t))^2,a_{min})
\end{aligned}
\right.
\end{equation} 
where $s^{\star}(t)$ is the current desired longitudinal distance gap, $v(t)$ the  longitudinal speed, $dv(t)$ the longitudinal speed difference with the lead, $s(t)$ the current distance gap. If there is no leading vehicle, $dv(t)$ and $1/s(t)$ are both set as 0. The IDM parameters are longitudinal desired speed $v_0$, time headway $T$, minimum gap $s_0$, acceleration coefficients $a_a$ and $a_b$, respectively. 

\begin{table}[htbp]
  \centering
  \caption{IDM parameters based on car-following trajectories in highD~\cite{kurtc2020studying}. }
    \begin{tabular}{cccccc}
    \toprule
    Parameters & \multicolumn{1}{c}{Desired speed $v_0$} & \multicolumn{1}{c}{Time headway $T$} & \multicolumn{1}{c}{Minimum gap $s_0$} & \multicolumn{1}{c}{Acceleration $a_a$} & \multicolumn{1}{c}{Deceleration $a_b$} \\
    \midrule
    Values & $v_e$/36.1 m/s & 0.8 s & 6.0 m & 1.0 m/s$^2$ & 1.0 m/s$^2$ \\
    \bottomrule
    \end{tabular}%
  \label{tab:IDM}%
\end{table}%

These IDM parameters are adopted from~\cite{kurtc2020studying} and summarized in Table~\ref{tab:IDM} in line with car-following trajectories in highD. The ego vehicle has a desired speed as its initial longitudinal speed $v_e$, and the desired speed of the surrounding is set as 36.1 m/s (130 km/h). The two vehicles share the same values of the remaining IDM parameters. As for the lateral cut-in behaviors, we adopt the polynomial curves in~\cite{mullakkal2020hybrid}, which can provide lateral accelerations with smooth trajectories.

\begin{figure}[htbp]
	\centering
	\subfigure[HSRS]{
		\includegraphics[width=0.7\textwidth]{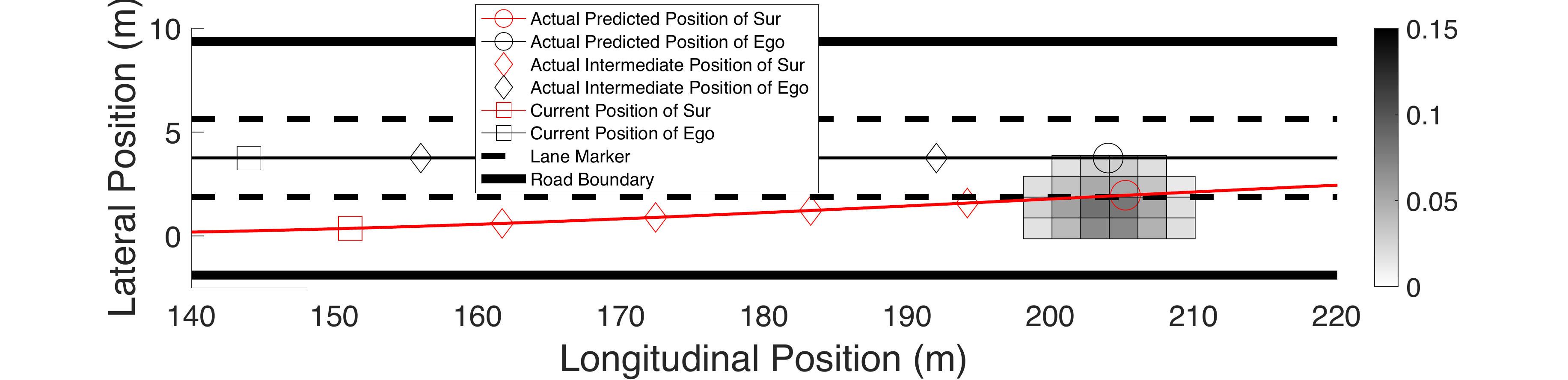}
	}
	\quad
		\subfigure[PSRS]{
		\includegraphics[width=0.7\textwidth]{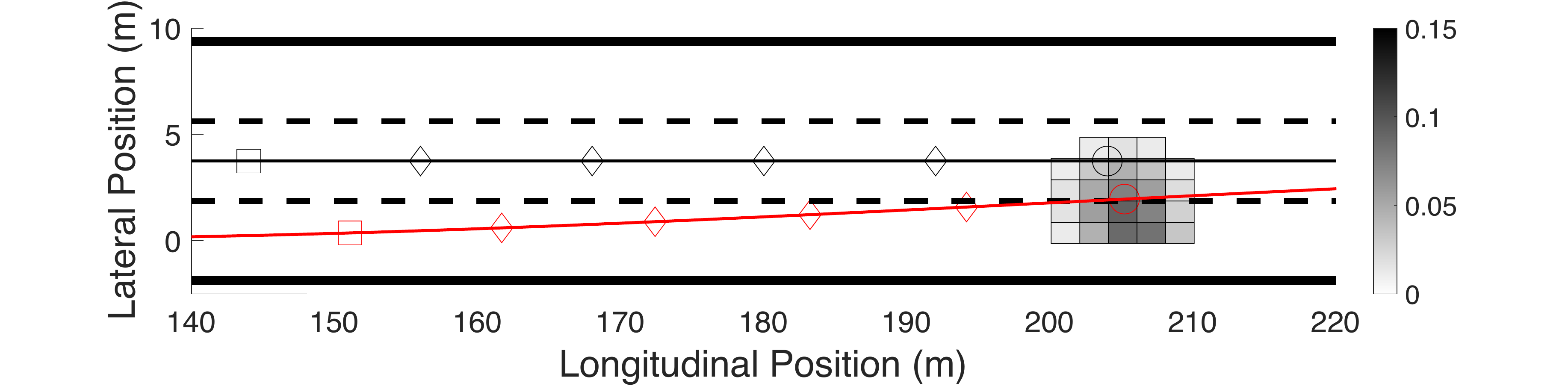}
	}
		\quad
	\subfigure[PSRS-3$\beta$]{
		\includegraphics[width=0.7\textwidth]{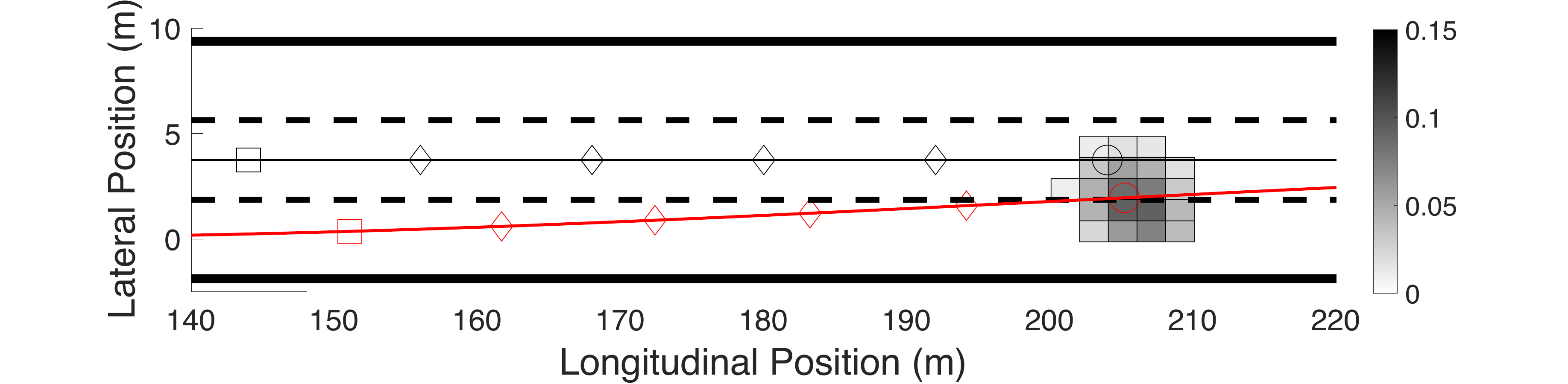}
	}
		\quad
	\subfigure[PSRS-5$\beta$]{
		\includegraphics[width=0.7\textwidth]{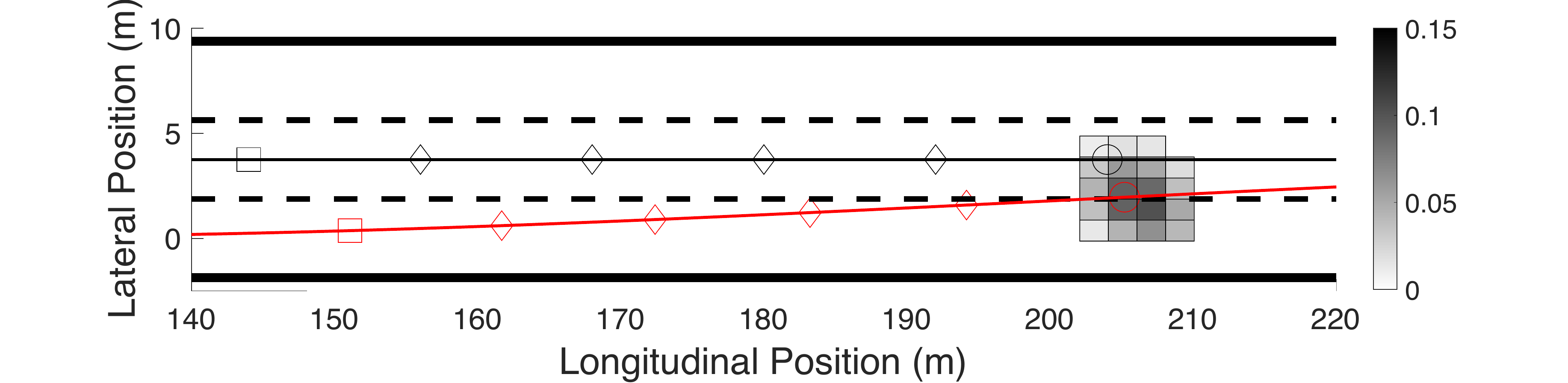}
	}
\caption{State probability distributions of  the stochastic FRS using different approaches in the simulated cut-in event (the current time $t=2.4$ seconds). Only position states with $>$ 0.01 probability are displayed.}\label{fig:SFRS_sim}
\end{figure}

\begin{figure}[htbp]
	\centering
	\subfigure[PSRS-3$\beta$]{
		\includegraphics[width=0.45\textwidth]{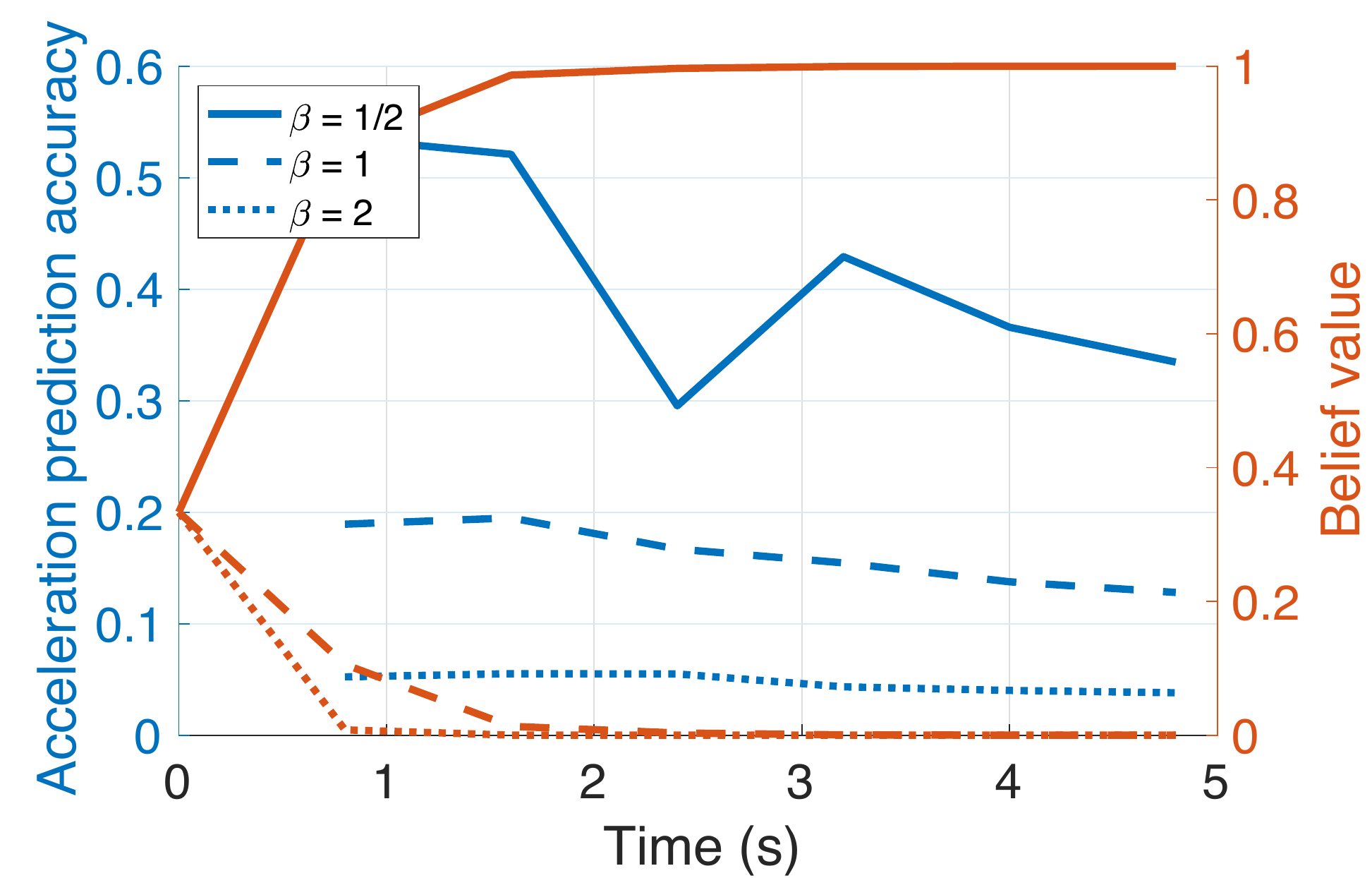}
	}
	\quad
		\subfigure[PSRS-5$\beta$]{
		\includegraphics[width=0.45\textwidth]{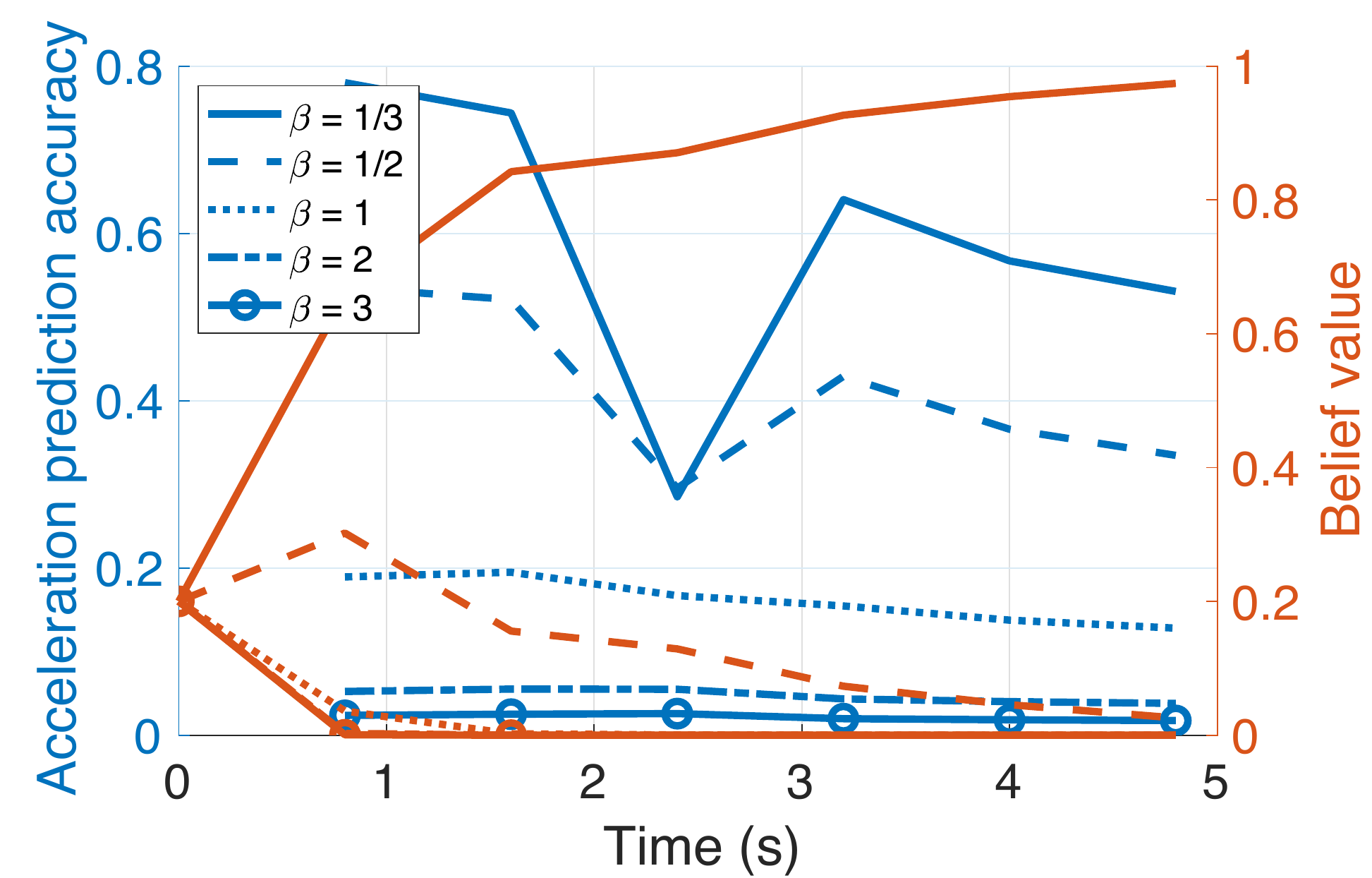}
	}
\caption{\blue{Belief dynamic changes for the simulated cut-in event leading to a crash at $t=5.0$ seconds. Blue lines represent acceleration prediction accuracy change with different coefficient settings, while orange lines indicate corresponding dynamic belief values.}}\label{fig:belief_update_sim}
\end{figure}

Given such cut-in scenario settings, a safety-critical cut-in event is created, and the crash occurs at around $t=5.0$ seconds. As seen in Fig.~\ref{fig:SFRS_sim}, the space state probabilities of the stochastic FRS are more concentrated using PSRS-3$\beta$ and PSRS-5$\beta$.  The future position prediction accuracy at the current time is 31.81\%, 33.4\%, 36.73\%, and 37.39\% using HSRS, PSRS, PSRS-3$\beta$ and PSRS-5$\beta$, respectively. The belief dynamic updates in the simulated cut-in event are illustrated in Fig.~\ref{fig:belief_update_sim}. 
\blue{By employing PSRS-3$\beta$ in Fig.~\ref{fig:belief_update_sim}(a), the belief value of the lowest $\beta$ shortly converges to one in 2 seconds. While in Fig.~\ref{fig:belief_update_sim}(b), the acceleration prediction accuracy based on confidence levels $\beta=1/3$ and $\beta=1/2$ drops to the same low value at $t=2.4$ seconds. Then the difference in  acceleration prediction accuracy becomes evident between $\beta = 1/3$ and $\beta=1/2$, leading to a  belief value convergence for $\beta=1/3$ later. Different confidence level settings and the dynamic belief value changes eventually lead to a better performance of PSRS-5$\beta$ compared to PSRS-3$\beta$.}

\begin{figure}[htb]
	\begin{center}
		\includegraphics[width=0.45\textwidth]{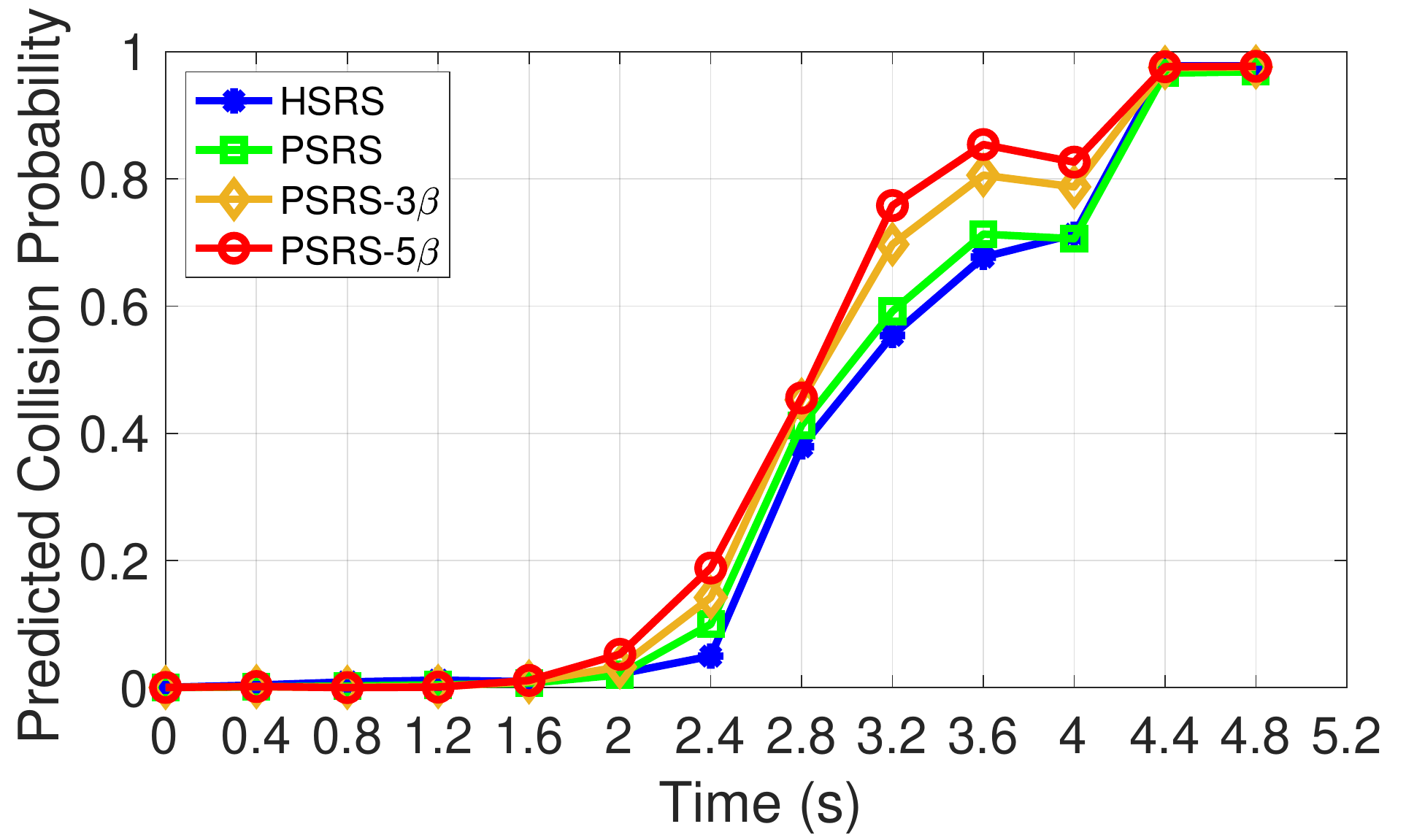}
		\caption{\blue{Collision probability estimation for a  cut-in crash event using different approaches. The initial longitudinal speeds of the ego and surrounding vehicles are 30 and 25 m/s, respectively. All four approaches successfully detect the  crash  at $t=5.0$ seconds. The estimated collision probability by PSRS-5$\beta$ has the most agile and effective performance thanks to  the employed confidence-aware stochastic FRS.}}\label{fig:col_compare}
	\end{center}
\end{figure}

\blue{The collision detection results are illustrated} in Fig.~\ref{fig:col_compare}. In the beginning (from 0 to 1.2 seconds), the four FRS-based approaches all predict a tiny collision probability, since the surrounding vehicle does not start lane change until $t=1$ second. After that, significant differences in the collision estimation are observed for the four approaches. For instance, when the estimated collision probability using PSRS-5$\beta$ is close to 0.20 at time $t=2.4$ seconds, PSRS without infusing confidence awareness predicts the collision probability as 0.10, and HSRS has a predicted collision probability about 0.05. It indicates that the stochastic FRS using confidence-aware prediction is agile and effective to identify potential collisions in the risky cut-in event. 

\begin{figure}[htbp]
	\centering
	\subfigure[PSRS-3$\beta$]{
		\includegraphics[width=0.45\textwidth]{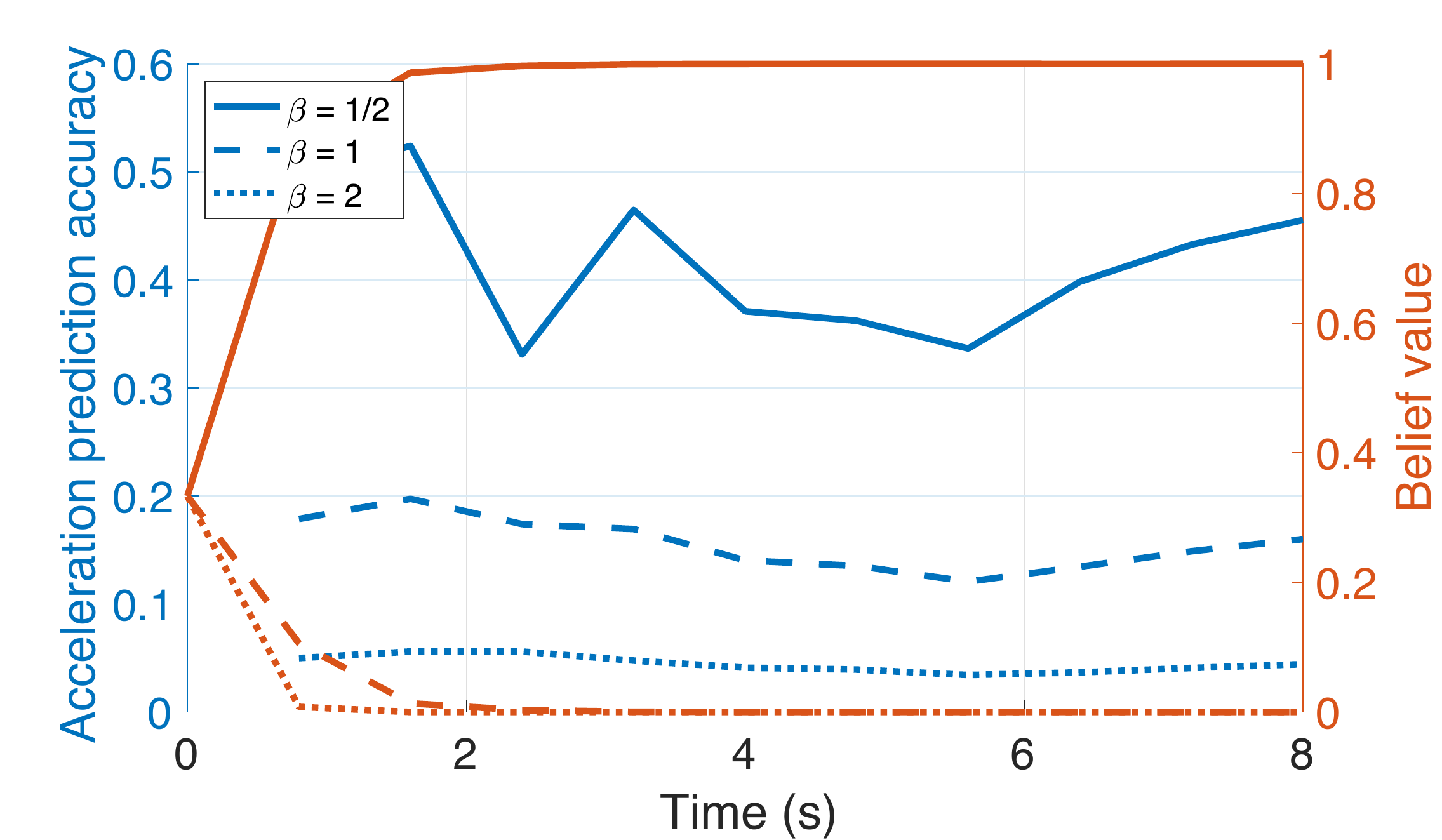}
	}
	\quad
		\subfigure[PSRS-5$\beta$]{
		\includegraphics[width=0.45\textwidth]{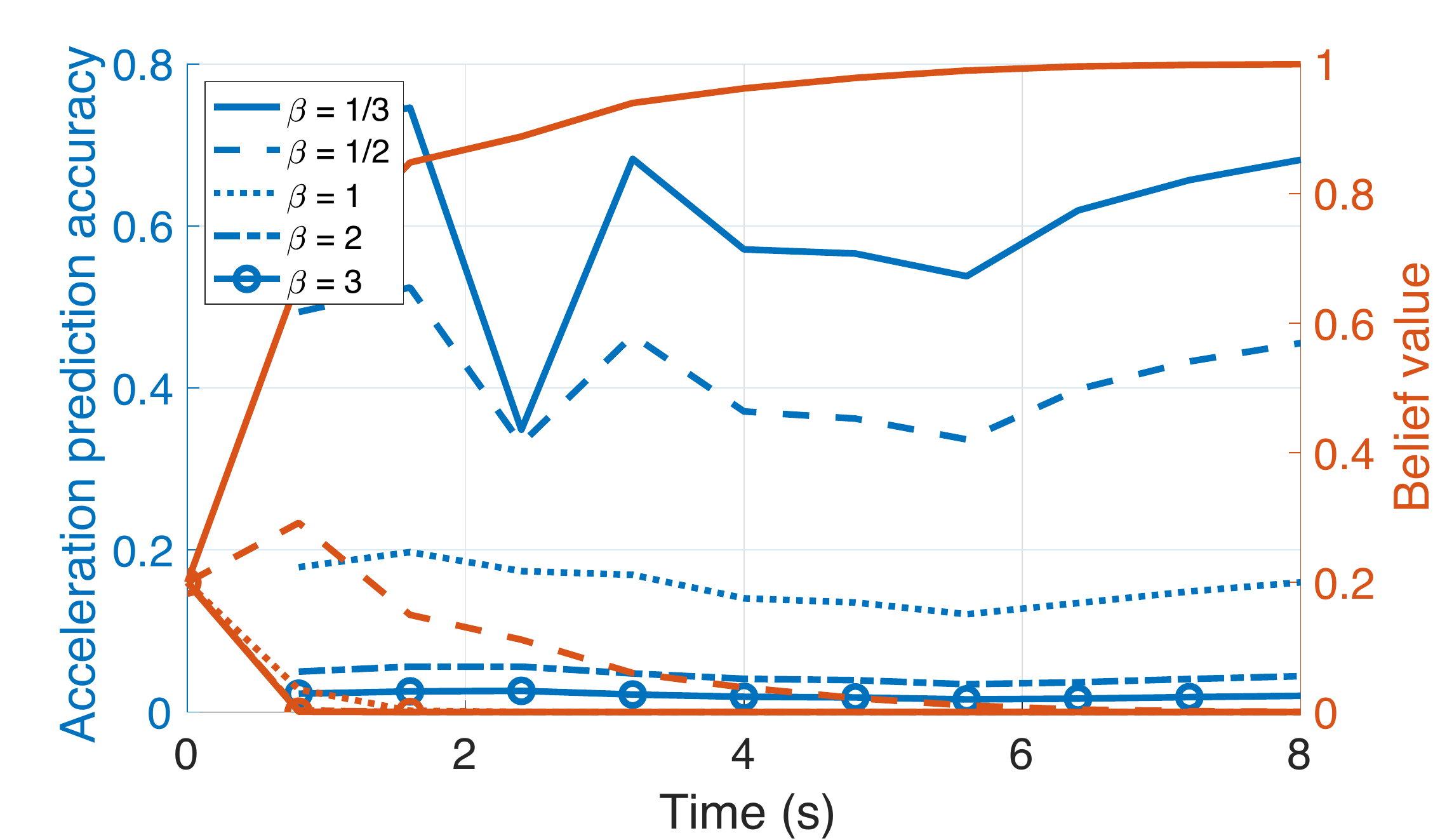}
	}
\caption{\blue{Belief dynamic changes for the simulated safety-critical cut-in event that does not lead to a crash. Blue lines represent acceleration prediction accuracy change with different coefficient settings, while orange lines indicate corresponding dynamic belief values.}}\label{fig:belief_update_sim_added}
\end{figure}

\begin{figure}[htb]
	\begin{center}
		\includegraphics[width=0.5\textwidth]{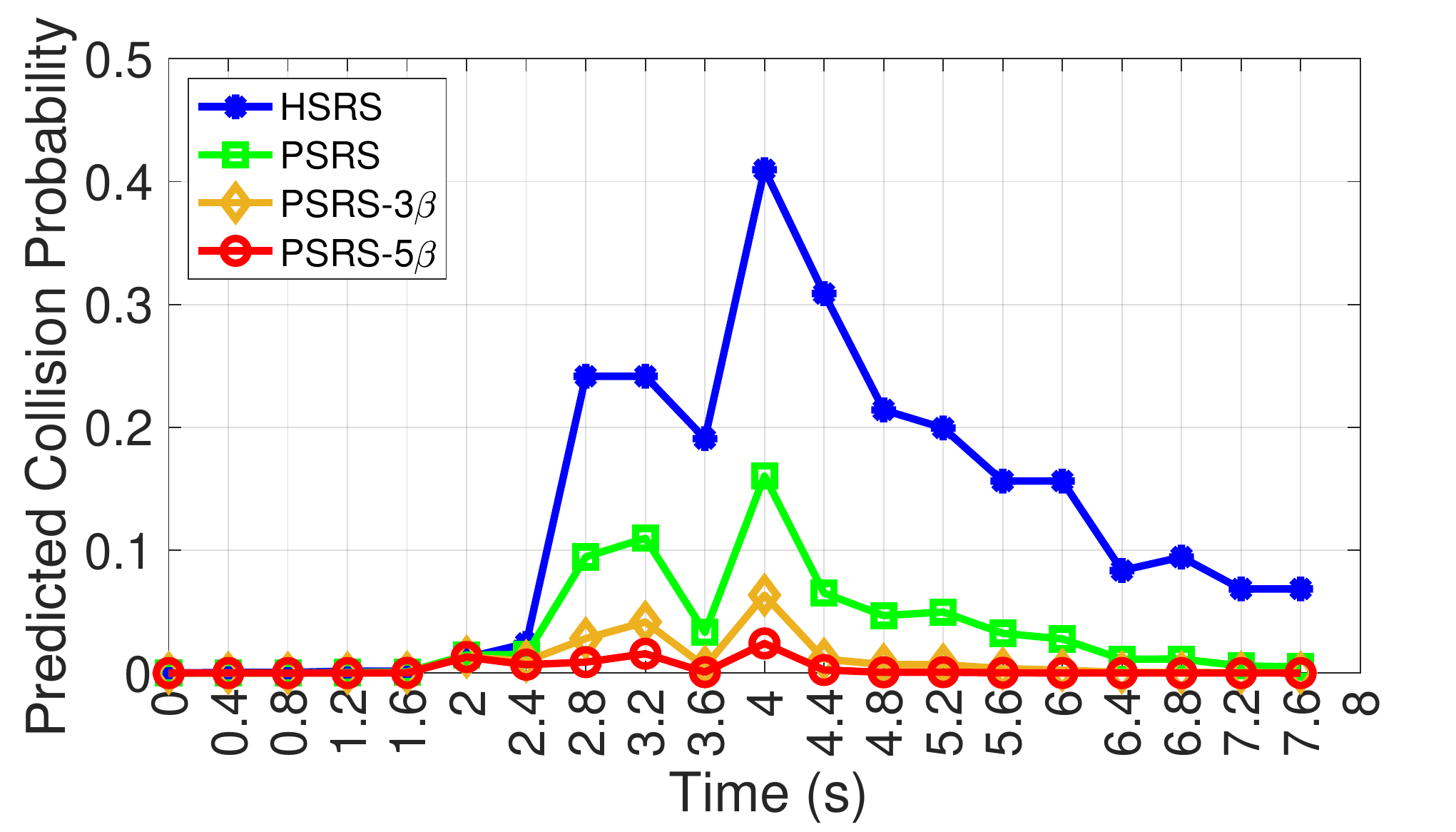}
		\caption{ \blue{Collision probability estimation for a safety-critical but collision-free cut-in event using different approaches.} 
  \blue{The initial longitudinal speeds of the ego and surrounding vehicles are 30 and 27 m/s, respectively. All four approaches  obtain a maximum collision probability at $t=4.0$ seconds when the cut-in vehicle is about to cross the lane marker. The estimated collision probability by PSRS-5$\beta$ is always below the pre-defined risk threshold 0.05.}}\label{fig:col_compare_added}
	\end{center}
\end{figure}

\blue{We also simulate a safety-critical cut-in event that does not lead to a crash, by changing the initial longitudinal speed of the surrounding vehicle as 27 m/s. The increased longitudinal velocity of the cut-in vehicle results in a larger longitudinal gap when it enters the target lane, thus leading to a safety-critical but collision-free cut-in event.  During the cut-in process, the belief dynamic changes shown in  Fig.~\ref{fig:belief_update_sim_added} have a very similar trend compared to that in the cut-in crash event, since the two events have the same parameter settings except for the longitudinal speed difference. As for the predicted collision probability illustrated in Fig.~\ref{fig:col_compare_added}, the heuristic based reachable set approach HSRS predicts the highest collision probability compared to the proposed prediction based approach PSRS. Specifically, HSRS reaches a maximum collision probability above 0.4 at $t=4$ seconds, which corresponds to the time that the surrounding cut-in vehicle crosses the lane marker. This is because HSRS cannot accurately predict the future positions of the surrounding vehicle, and incorrectly estimates a high collision probability between two vehicles. The simplified heuristic approach HSRS also calculates a collision probability 0.07 when the surrounding vehicle completes the cut-in process and becomes the leader of the ego vehicle in the target lane. Among the three PSRS approaches with different confidence aware coefficients, we observe a maximum collision probability is also obtained at the critical time $t=4$ seconds. Without the infusion of confidence awareness, PSRS calculates a collision probability of 0.16, while PSRS-3$\beta$/-5$\beta$ obtains a smaller collision probability 0.07/0.03, thanks to a more concentrated stochastic FRS. Given the simulated cut-in event is safety-critical but collision-free, the PSRS approaches indeed accurately captured the potential collision risk with a smaller collision probability compared to that of HSRS. In particular, the estimated collision probability by  PSRS-5$\beta$ is always below the pre-defined threshold 0.05, indicating that the confidence-infused prediction approach PSRS-5$\beta$ cannot only accurately identify risky cut-in crash, but also avoid false positive results in a safety-critical but collision-free event.}

\begin{figure}[!htb]
	\begin{center}
		\includegraphics[width=0.5\textwidth]{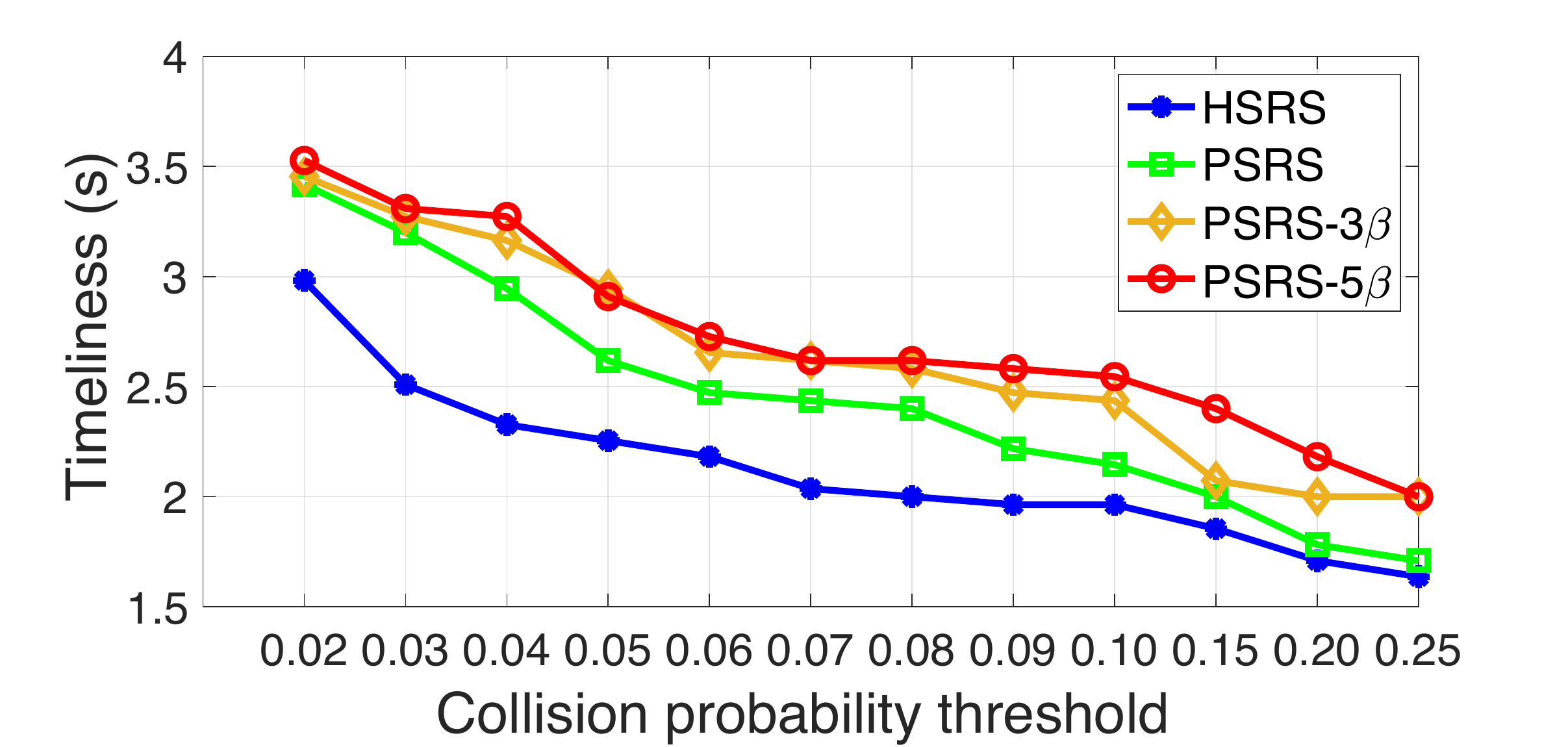}
		\caption{Average timeliness performance under different collision thresholds. The timeliness value represents the time that remains to the crash when the estimated collision probability has reached a threshold.}\label{fig:col_statistic}
	\end{center}
\end{figure}

Under the same simulated cut-in scenario settings, we further vary the initial longitudinal speeds of the two vehicles from 20 m/s to 35 m/s with an increment 1 m/s, leading to $16\times 16 = 256$ cut-in events. Based on the employed longitudinal and lateral driving behavior models, 33 crashes are identified when $4 \leq v_{e} - v_{s} \leq 6$ m/s. To compare the collision detection performance,  we statistically analyze the simulated 21 cut-in events where a crash occurs. The analysis results are illustrated in Fig.~\ref{fig:col_statistic}, where the timeliness value represents the average time that remains to the crash when the collision probability has reached a threshold. No matter the selection of the collision probability threshold, our proposed approaches infusing confidence awareness (i.e., PSRS-5$\beta$/PSRS-3$\beta$) can achieve a larger timeliness value, indicating a more adequate reaction time to potential crashes. The selection of a suitable collision probability threshold could vary in different scenarios and we leave it for future research.

\subsection{The integrated collision detection framework}

Based on the BRS and stochastic FRS, we propose an integrated  collision detection framework. We first provide comparative results between two reachable set techniques, i.e., BRS and FRS. This explains the selection of BRS rather FRS to formally check driving safety in the first step of the framework.

\begin{figure}[!htbp]
	\centering
	\subfigure[]{
		\includegraphics[width=0.40\textwidth]{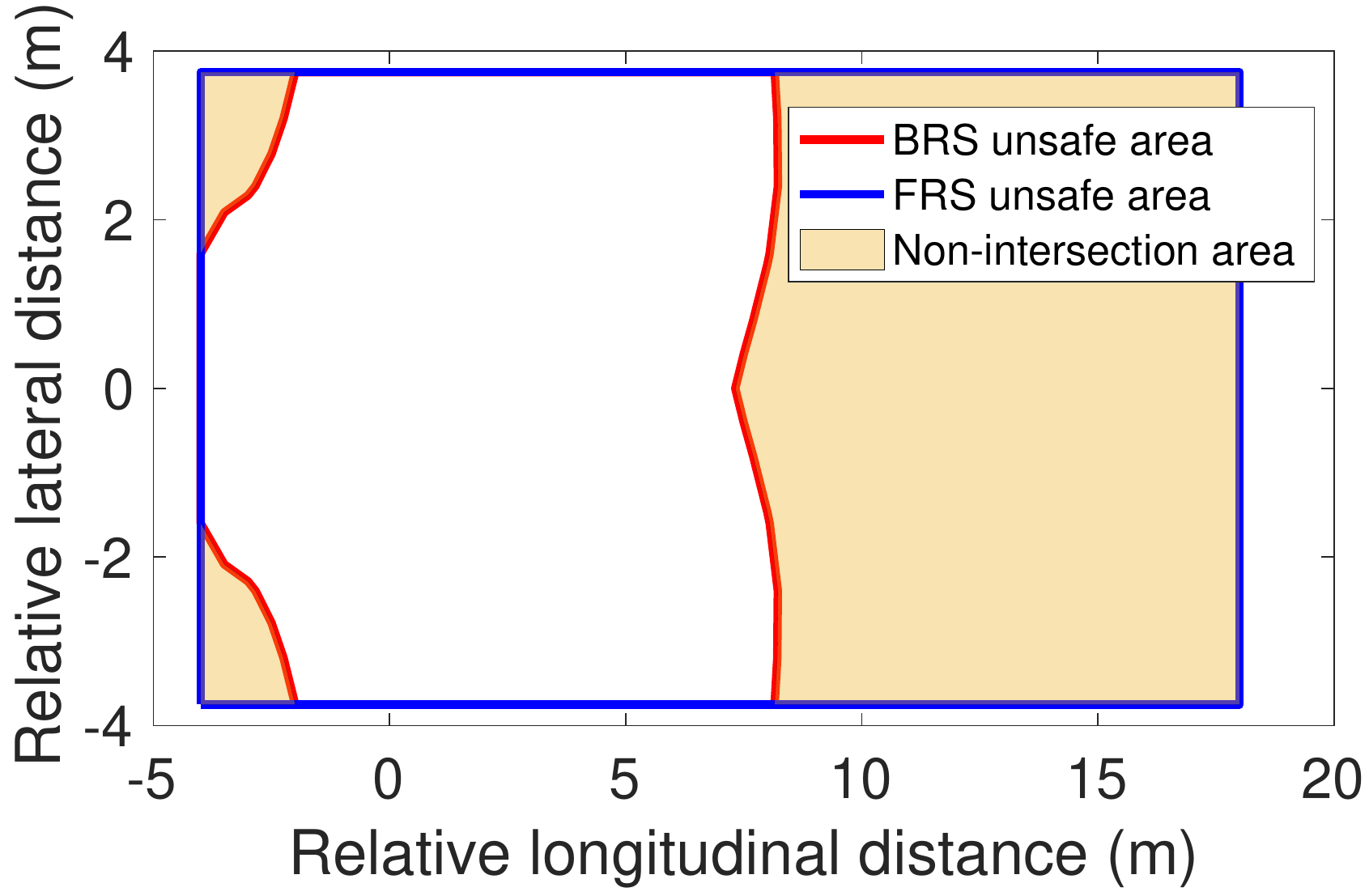}
	}
	\quad
		\subfigure[]{
		\includegraphics[width=0.40\textwidth]{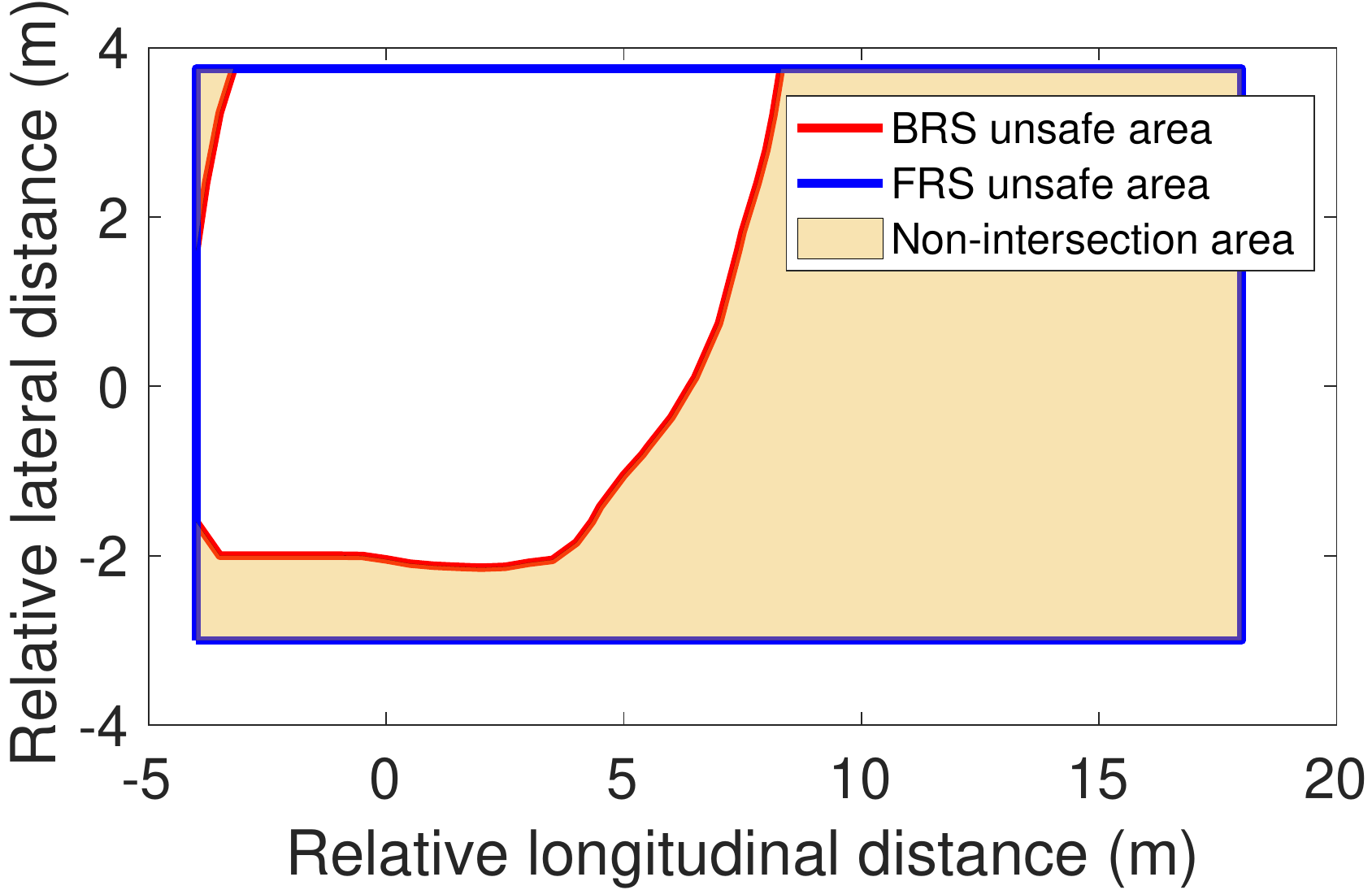}
	}
\caption{Comparisons for the BRS and FRS based unsafe areas. 28 m/s and 30 m/s for the surrounding and ego vehicles.  The ego is in the origin. (a) Lateral speed of the surrounding is 0 m/s. The relative heading angle is 0 degree. (b) Lateral speed of the surrounding is -1 m/s. The relative heading angle range is from 9 to 45 degrees. }\label{fig:BRSvsFRS}
\end{figure}

As shown in Fig.~\ref{fig:BRSvsFRS}, we set the longitudinal speed as 30 and 28 m/s for the ego and the surrounding vehicle respectively, and illustrate comparative results using both the BRS and FRS as an example. The BRS is directly employed to identify unsafe area, once its cached state value function is less than zero.  To obtain the unsafe area identified by the FRS, we first need to enumerate initial relative surrounding vehicle positions, and check whether two vehicles could collide using the FRS with a prediction horizon. Then we enclose all relative positions that could lead to a crash and identify the enclosed area in blue line as unsafe. Note that we limit the unsafe area between -3.75 to 3.75 meters in the lateral direction, since we only need to address potential risky interactions between adjacent lanes. Similarly, we  do not check relative longitudinal positions behind the ego.  As shown in Fig.~\ref{fig:BRSvsFRS}, the FRS indeed identifies a larger unsafe area compared to that using the BRS. This is reasonable, as the BRS further considers ego reaction to the surrounding vehicle, leading to a smaller unsafe area. Specifically, the identified unsafe areas are symmetric in Fig~\ref{fig:BRSvsFRS}(a), since the lateral speed and relative heading angles of the initial vehicle states are both zeros. In contrast, the surrounding vehicle has an initial lateral speed and a  relative heading angle in Fig~\ref{fig:BRSvsFRS}(b).  Then based on the FRS,  we  observe that the area with a lateral position between -3 to -3.75 meters is now safe due to the lateral speed of the surrounding vehicle. The BRS now has an asymmetric shape due to the lateral speed, and the area with minus lateral relative position becomes safer. This example verifies that  the BRS is indeed less conservative than  the FRS, since the unsafe area identified by the BRS is smaller than that by the FRS, and the BRS identified unsafe area is a subset of the FRS identified area. This observation is consistent with the definitions of BRS and FRS.

\begin{figure}[!htb]	
\begin{center}
\includegraphics[width=0.7\textwidth]{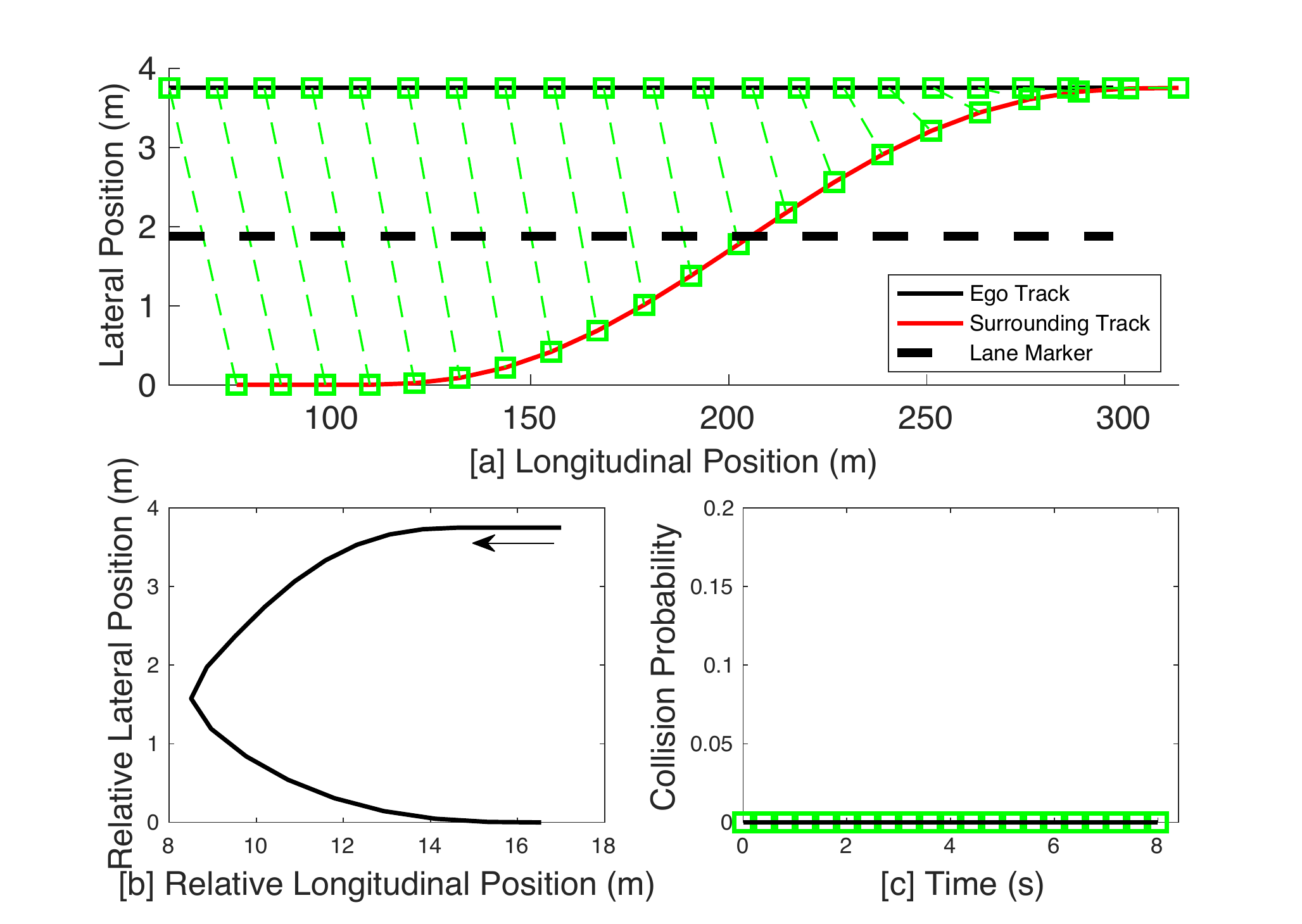}\caption{Collision detection results using the integrated framework. The initial longitudinal speeds of the ego and the surrounding are 30 and 28 m/s, respectively. In subfigure (a), the  squares denote vehicle positions at each time step and the two vehicle positions at the same step are connected by dash lines. Each connected position pair in green indicates that the car-car interaction is identified as theoretically safe by the BRS. Subfigure (b) illustrates the relative vehicle positions, where the arrow indicates the evolution direction. \blue{In subfigure (c), the estimated collision probability marked by  green squares remains zero}, since the relative positions between two vehicles are outside of the unsafe area identified by the BRS. }\label{fig:frame_safe_cutin}
	\end{center}
\end{figure}

Then we provide two specific cut-in event to validate the proposed  collision detection framework. The cut-in events are both selected from the simulated trajectories in Section~\ref{sec:sim_tracks}. In the first cut-in event, the initial longitudinal speeds of the ego and the surrounding are 30 and 28 m/s, respectively. This leads to non-risky and safe interactions between the two vehicles. As shown in Fig.~\ref{fig:frame_safe_cutin}, during the lane change, the integrated framework ensures safety, since the relative positions (see Fig.~\ref{fig:frame_safe_cutin}(a) and Fig.~\ref{fig:frame_safe_cutin}(b)) are not inside the unsafe area identified by the BRS. Thus, the establishment of stochastic FRS is not activated and the estimated collision probability always remains zero in Fig.~\ref{fig:frame_safe_cutin}(c).

\begin{figure}[h!]	
\begin{center}
\includegraphics[width=0.9\textwidth]{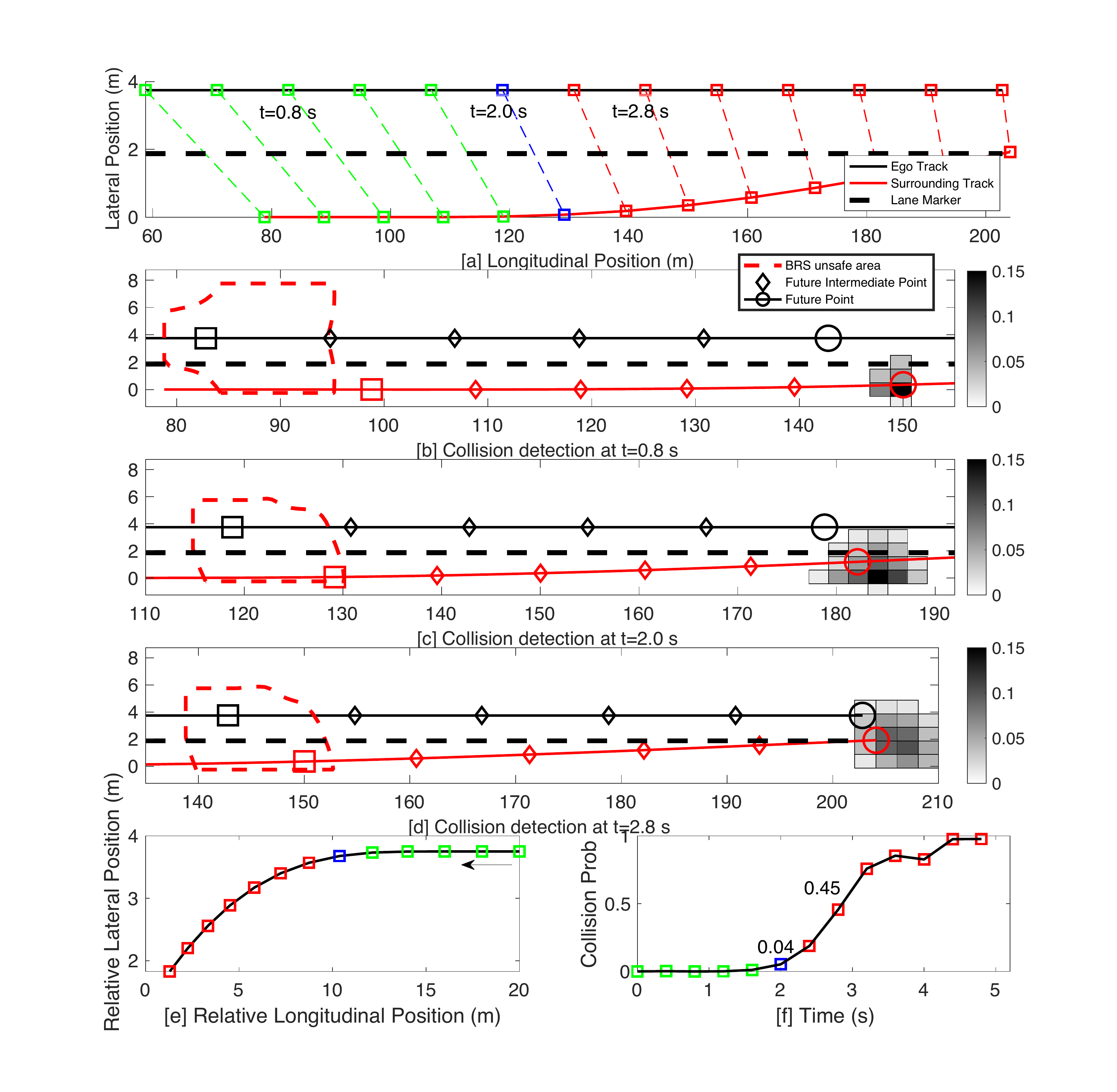}\caption{Collision detection using the integrated framework. The initial longitudinal speeds of the ego and the surrounding are 30 and 25 m/s, respectively. In subfigure (a), the  squares denote vehicle positions at each time step and the two vehicle positions at the same step are connected by dash lines. Each connected position pair in green indicates that the car-car interaction is identified as theoretically safe by the BRS, pair in blue indicates the interaction has an estimated collision probability below a threshold (0.05 in this case), and pairs in red indicate the interaction has a collision probability above the threshold. Subfigures (b) - (d) illustrate the shapes of BRS and FRS at t=0.8, 2.0 and 2.8 seconds respectively. Note that to calculate the collision probability, the stochastic FRS is established at each future time step including the four intermediate time points, and we only show the stochastic FRS at the last prediction time step for convenience. In subfigure (b), the current position of the surrounding vehicle is outside the BRS unsafe area of the ego, thus the stochastic FRS does not need to be constructed in practice. We just show the stochastic FRS at t=0.8 seconds in subfigure (b) for completeness. Subfigure (e) illustrates the relative vehicle positions, where the arrow indicates the evolution direction. The estimated collision probability in subfigure (f) is marked as green at the beginning, then blue  and red in the end phase. }\label{fig:frame_risk_cutin}
	\end{center}
\end{figure}



In the second cut-in event, the initial longitudinal speeds of the ego and the surrounding are 30 and 25 m/s, respectively, resulting in a collision about $t=5$  seconds shown in Fig.~\ref{fig:frame_risk_cutin}. When we use the proposed collision detection framework in this safety-critical event, safety can be ensured at the beginning (see green squares in Fig.~\ref{fig:frame_risk_cutin}). This is because the surrounding vehicle starts lane-change maneuver at  $t=1.0$ second, and driving risk can be only captured until the surrounding vehicle has clear lane-change intention.  Thus at $t=2.0$ seconds, the proposed framework can no longer ensure driving safety (see Fig.~\ref{fig:frame_risk_cutin}(c) where the surrounding vehicle is inside the BRS unsafe area of the ego), and the stochastic FRS is established to estimate a collision probability accurately. However, the obtained collision probability is below a pre-defined threshold 0.05 at $t=2.0$ seconds, and we mark it in blue. Afterwards, the framework provides an estimated collision probability above the threshold until collision happens at $t=5.0$ seconds, where the estimated collision gradually increases to one (see Fig.~\ref{fig:frame_risk_cutin}(f)). In summary, the proposed framework can reasonably ensure  safety at the beginning, and then effectively identify the potential crash thanks to the established stochastic FRS. 

\blue{In terms of computational efficiency, the proposed framework is promising for real-time risk assessment applications, and combining BRS can lead to further time savings. This is due to that the BRS cached in a look-up table can quickly identify if a vehicle is theoretically safe, taking only a few milliseconds, whereas the stochastic FRS risk assessment can take between 30 to 50 milliseconds given the experimental settings. As most highway driving situations do not involve safety-critical vehicle interactions~\cite{krajewski2018highd}, BRS can prevent unnecessary computation time for stochastic FRS construction if safety is ensured. Note that the real-time capability is realized based on an advanced GPU enabling 2048$\times$28 parallel computation with power consumption up to 1.5 kWh~\cite{mittal2014survey}. If the GPU runs for stochastic FRS construction all the time, it could occupy 10\% of the entire power consumption of an electric vehicle. However, naturalistic driving data has shown that vehicle interactions are not safety-critical in most instances~\cite{krajewski2018highd}. It implies that safety can be ensured by BRS in most cases. Consequently, the integrated risk assessment framework can also benefit from significant power consumption reduction by not using GPU computation in most instances.}

\section{Conclusions} \label{sec:conc}

  We have developed a reachability-based framework for collision detection in highway driving. \blue{Inspired by two different reachability analysis approaches, a cached backward reachable set is firstly employed to formally verify whether the current interaction safety can be theoretically ensured. Otherwise, a prediction-based confidence-aware stochastic forward reachable set (FRS) is calculated online at each predicted time step for collision probability estimation. If the estimated collision probability exceeds a pre-defined threshold, the ego vehicle can then execute brakes or swerve to avoid potential crashes. In doing so, the proposed framework can ensure risk-free interactions in non-risky events and provide an accurate collision estimation in safety-critical events.} 
        
        \blue{Working with both naturalistic driving data and simulated safety-critical cut-in events, we conducted extensive experiments to validate the performance of the proposed acceleration prediction model, which effectively considered interactions between vehicles for constructing a stochastic FRS. Furthermore, we analyzed and discussed how infusing confidence belief can improve acceleration prediction accuracy. Simulation results have shown that the proposed FRS-based approach PSRS-5$\beta$ with five confidence levels  is capable of adjusting complicated highway scenarios, leading to more agile collision detection results. The integrated framework has also been tested in both risky and non-risky events. We have demonstrated that the proposed framework cannot only accurately identify potential collision events, but also have the potential to avoid false positive results in safety-critical but collision-free events.} 

The proposed approach assumes the use of an advanced GPU for efficient parallel computing on the stochastic FRS. 
This requires further efforts to conduct hardware implications \blue{with lower power consumption} and real-world testing. Besides, the developed acceleration predictor is trained on highD dataset, which only contains highway trajectories. We can only apply our approach to diverse scenarios, e.g., urban and country-road environments, by training the predictor with a more diverse dataset.  Future works could integrate the risk assessment to enable safer and more efficient motion planning, and employ the proposed risk assessment framework on an actual vehicle.

\section*{Appendix}
\subsection*{Equivalent transformation of vehicle dynamics between FRS and BRS}

In the overall framework,  different  vehicle models are considered in the FRS and BRS. In this appendix, the transfer of the vehicle dynamics between the FRS and BRS is detailed, which is used to map the range of state-action space (i.e., the control input) between the two reachable sets. Note that in this appendix, we use more common notations to redefine the system dynamics and coordinates for an easier understanding of the equivalent transformation.   

According to Section~\ref{sec:dynamics}, the point mass vehicle model is applied to describe the motion of the  vehicle in the FRS. 
\begin{subequations}
\begin{align}
    \ddot{y}&=a_{y}\\
\ddot{x}&=a_{x}\\
\dot{Y}&=\dot{x} \sin \psi+\dot{y} \cos \psi\\
\dot{X}&=\dot{x} \cos \psi-\dot{y} \sin \psi
\end{align}
\end{subequations}
where $\dot{X}$ and $\dot{Y}$ are two-dimensional velocities in the global coordinate system, and $\psi \in (0,2\pi)$ is the yaw angle corresponding to the global coordinates.  $\dot{x}$ and $\dot{y}$ are longitude and lateral speed in vehicle coordinates. $a_{x}$ and $a_{y}$ are longitudinal and lateral accelerations. For the transformation of control input from the point mass model in the FRS to the ego vehicle's bicycle model in the BRS, we add the influence of angular velocity $\dot{\psi}$ by 
\begin{equation}
a_{y}=\dot{\psi} \dot{x}
\label{add_influence}
\end{equation}

Take the derivative of $\dot{Y}$ and $\dot{X}$ to obtain $\ddot{Y}$ and $\ddot{X}$:
\begin{equation}
    \begin{aligned}
\ddot{Y} &=\ddot{x} \sin \psi+(\dot{x} \cos \psi) \dot{\psi}+\ddot{y} \cos \psi-(\dot{y} \sin \psi) \dot{\psi} \\
&=\ddot{x} \sin \psi+a_{y} \cos \psi+\ddot{y} \cos \psi-\frac{y}{\dot{x}} a_{y} \sin \psi \\
&=a_{x} \sin \psi+2 a_{y} \cos \psi-\frac{\dot{y}}{\dot{x}} a_{y} \sin \psi
\end{aligned}
\end{equation}

\begin{equation}
    \begin{aligned}
\ddot{X} &=\ddot{x} \cos \psi-(\dot{x} \sin \psi) \dot{\psi}-\ddot{y} \sin \psi-(\dot{y} \cos \psi) \dot{\psi} \\
&=\ddot{x} \cos \psi-a_{y} \sin \psi-\ddot{y} \sin \psi-\frac{y}{\dot{x}} a_{y} \cos \psi \\
&=a_{x} \cos \psi-2 a_{y} \sin \psi-\frac{\dot{y}}{\dot{x}} a_{y} \cos \psi
\end{aligned}
\end{equation}
Finally, the acceleration and velocity in the longitudinal and lateral are formulated by $\dot{X}$ and $\dot{Y}$:
\begin{subequations}
\begin{align}
a_{x}&=\frac{\dot{y}}{\dot{x}} a_{y}+(\ddot{X} \sin \psi-\ddot{Y} \cos \psi) \label{ax}\\
a_{y}&=\frac{1}{2}(\ddot{Y} \cos \psi-\ddot{X} \sin \psi) \label{ay}\\
v_{x}&=\dot{x} \\
v_{y}&=\dot{y}
\end{align}
\end{subequations}
Meanwhile, considering the ego vehicle model in the BRS:
\begin{subequations}
\begin{align}
\dot{x} &=v \cos \left(\psi+\beta\right) \\
\dot{y} &=v \sin \left(\psi+\beta\right) \\
\dot{v} &=a \\
\dot{\psi} &=\frac{v_{}}{l_{r}} \sin \left(\beta\right) \\
\beta &=\tan ^{-1}\left(\frac{l_{r}}{l_{f}+l_{r}} \tan  \left(\delta_{f}\right)\right) \label{beta}
\end{align}
\end{subequations}
where $l_{r}=1.738$ m and $l_{f}=1.058$ m are rear and front wheelbase. $v =\sqrt{v_{x}^2+v_{y}^2}$ and $a =\sqrt{a_{x}^2+a_{y}^2}$ is the synthesis velocity and acceleration, respectively. $\beta$ is the slip angle and $\delta_{f}$ is the steering angle. For the ego vehicle model (bicycle model) in the BRS, the range of vehicle control input $\beta$ and $a_{ego}$ can be calculated by the scale of $\ddot{X}$, $\ddot{Y}$ and $\psi$ based on \eqref{add_influence}, \eqref{ax}, \eqref{ay} and \eqref{beta}. In this work, the range of $\ddot{X}$ and $\ddot{Y}$ are (-5,3) m/s$^2$ and (-1.5,1.5) m/s$^2$. And the range of $\dot{x}$ and $\dot{y}$ are (20,40) m/s and (-1.5,1.5) m/s, respectively. Considering that the surrounding vehicle model in the BRS is the simplified unicycle model, the range of disturbance $a_{sur}$ is the same as $a_{ego}$ in the ego vehicle model. Meanwhile, the range of disturbance angular acceleration $\omega_{sur} = \dot{\psi}$ can be calculated by \eqref{add_influence}, \eqref{ax} and \eqref{ay}. Finally, we can obtain the equivalent range of control input ($\beta$, $a_{ego}$) and disturbance ($a_{sur}$, $\omega_{sur}$).

\bibliographystyle{IEEEtran} 
\bibliography{RA}{}

\begin{thebibliography}{10}
\providecommand{\url}[1]{#1}
\csname url@samestyle\endcsname
\providecommand{\newblock}{\relax}
\providecommand{\bibinfo}[2]{#2}
\providecommand{\BIBentrySTDinterwordspacing}{\spaceskip=0pt\relax}
\providecommand{\BIBentryALTinterwordstretchfactor}{4}
\providecommand{\BIBentryALTinterwordspacing}{\spaceskip=\fontdimen2\font plus
\BIBentryALTinterwordstretchfactor\fontdimen3\font minus
  \fontdimen4\font\relax}
\providecommand{\BIBforeignlanguage}[2]{{%
\expandafter\ifx\csname l@#1\endcsname\relax
\typeout{** WARNING: IEEEtran.bst: No hyphenation pattern has been}%
\typeout{** loaded for the language `#1'. Using the pattern for}%
\typeout{** the default language instead.}%
\else
\language=\csname l@#1\endcsname
\fi
#2}}
\providecommand{\BIBdecl}{\relax}
\BIBdecl

\bibitem{kalra2016driving}
N.~Kalra and S.~M. Paddock, ``Driving to safety: How many miles of driving
  would it take to demonstrate autonomous vehicle reliability?''
  \emph{Transportation Research Part A: Policy and Practice}, vol.~94, pp.
  182--193, 2016.

\bibitem{lefevre2014survey}
S.~Lef{\`e}vre, D.~Vasquez, and C.~Laugier, ``A survey on motion prediction and
  risk assessment for intelligent vehicles,'' \emph{ROBOMECH journal}, vol.~1,
  no.~1, pp. 1--14, 2014.

\bibitem{mukhtar_vehicle_2015}
A.~Mukhtar, L.~Xia, and T.~B. Tang, ``Vehicle {Detection} {Techniques} for
  {Collision} {Avoidance} {Systems}: {A} {Review},'' \emph{IEEE Transactions on
  Intelligent Transportation Systems}, vol.~16, no.~5, pp. 2318--2338, 2015.

\bibitem{pek2020using}
C.~Pek, S.~Manzinger, M.~Koschi, and M.~Althoff, ``Using online verification to
  prevent autonomous vehicles from causing accidents,'' \emph{Nature Machine
  Intelligence}, vol.~2, no.~9, pp. 518--528, 2020.

\bibitem{falcone2011predictive}
P.~Falcone, M.~Ali, and J.~Sjoberg, ``Predictive threat assessment via
  reachability analysis and set invariance theory,'' \emph{IEEE Transactions on
  Intelligent Transportation Systems}, vol.~12, no.~4, pp. 1352--1361, 2011.

\bibitem{althoff2014online}
M.~Althoff and J.~M. Dolan, ``Online verification of automated road vehicles
  using reachability analysis,'' \emph{IEEE Transactions on Robotics}, vol.~30,
  no.~4, pp. 903--918, 2014.

\bibitem{althoff2021set}
M.~Althoff, G.~Frehse, and A.~Girard, ``Set propagation techniques for
  reachability analysis,'' \emph{Annual Review of Control, Robotics, and
  Autonomous Systems}, vol.~4, pp. 369--395, 2021.

\bibitem{leung2020infusing}
K.~Leung, E.~Schmerling, M.~Zhang, M.~Chen, J.~Talbot, J.~C. Gerdes, and
  M.~Pavone, ``On infusing reachability-based safety assurance within planning
  frameworks for human--robot vehicle interactions,'' \emph{The International
  Journal of Robotics Research}, vol.~39, no. 10-11, pp. 1326--1345, 2020.

\bibitem{chapman2021risk}
M.~P. Chapman, R.~Bonalli, K.~M. Smith, I.~Yang, M.~Pavone, and C.~J. Tomlin,
  ``Risk-sensitive safety analysis using conditional value-at-risk,''
  \emph{arXiv preprint arXiv:2101.12086}, 2021.

\bibitem{lee1976theory}
D.~N. Lee, ``A theory of visual control of braking based on information about
  time-to-collision,'' \emph{Perception}, vol.~5, no.~4, pp. 437--459, 1976.

\bibitem{minderhoud2001extended}
M.~M. Minderhoud and P.~H. Bovy, ``Extended time-to-collision measures for road
  traffic safety assessment,'' \emph{Accident Analysis \& Prevention}, vol.~33,
  no.~1, pp. 89--97, 2001.

\bibitem{vogel2003comparison}
K.~Vogel, ``A comparison of headway and time to collision as safety
  indicators,'' \emph{Accident Analysis \& Prevention}, vol.~35, no.~3, pp.
  427--433, 2003.

\bibitem{mammar2006time}
S.~Mammar, S.~Glaser, and M.~Netto, ``Time to line crossing for lane departure
  avoidance: A theoretical study and an experimental setting,'' \emph{IEEE
  Transactions on Intelligent Transportation Systems}, vol.~7, no.~2, pp.
  226--241, 2006.

\bibitem{saunier2008probabilistic}
N.~Saunier and T.~Sayed, ``Probabilistic framework for automated analysis of
  exposure to road collisions,'' \emph{Transportation Research Record}, vol.
  2083, no.~1, pp. 96--104, 2008.

\bibitem{davis2011outline}
G.~A. Davis, J.~Hourdos, H.~Xiong, and I.~Chatterjee, ``Outline for a causal
  model of traffic conflicts and crashes,'' \emph{Accident Analysis \&
  Prevention}, vol.~43, no.~6, pp. 1907--1919, 2011.

\bibitem{kuang2015tree}
Y.~Kuang, X.~Qu, and S.~Wang, ``A tree-structured crash surrogate measure for
  freeways,'' \emph{Accident Analysis \& Prevention}, vol.~77, pp. 137--148,
  2015.

\bibitem{mullakkal2020probabilistic}
F.~A. Mullakkal-Babu, M.~Wang, X.~He, B.~van Arem, and R.~Happee,
  ``Probabilistic field approach for motorway driving risk assessment,''
  \emph{Transportation Research Part C: Emerging Technologies}, vol. 118, p.
  102716, 2020.

\bibitem{katare_embedded_2019}
D.~Katare and M.~El-Sharkawy, ``Embedded system enabled vehicle collision
  detection: an {ANN} classifier,'' in \emph{2019 {IEEE} 9th {Annual}
  {Computing} and {Communication} {Workshop} and {Conference} ({CCWC})}.\hskip
  1em plus 0.5em minus 0.4em\relax IEEE, 2019, pp. 0284--0289.

\bibitem{saleh_kangaroo_2016}
K.~Saleh, M.~Hossny, and S.~Nahavandi, ``Kangaroo vehicle collision detection
  using deep semantic segmentation convolutional neural network,'' in
  \emph{2016 {International} {Conference} on {Digital} {Image} {Computing}:
  {Techniques} and {Applications} ({DICTA})}.\hskip 1em plus 0.5em minus
  0.4em\relax IEEE, 2016, pp. 1--7.

\bibitem{laugier_probabilistic_2011}
C.~Laugier, I.~E. Paromtchik, M.~Perrollaz, M.~Yong, J.-D. Yoder, C.~Tay,
  K.~Mekhnacha, and A.~Nègre, ``Probabilistic {Analysis} of {Dynamic} {Scenes}
  and {Collision} {Risks} {Assessment} to {Improve} {Driving} {Safety},''
  \emph{IEEE Intelligent Transportation Systems Magazine}, vol.~3, no.~4, pp.
  4--19, 2011.

\bibitem{annell_probabilistic_2016}
S.~Annell, A.~Gratner, and L.~Svensson, ``Probabilistic collision estimation
  system for autonomous vehicles,'' in \emph{2016 {IEEE} 19th {International}
  {Conference} on {Intelligent} {Transportation} {Systems} ({ITSC})}, 2016, pp.
  473--478.

\bibitem{kim_collision_2018}
J.~Kim and D.~Kum, ``Collision {Risk} {Assessment} {Algorithm} via
  {Lane}-{Based} {Probabilistic} {Motion} {Prediction} of {Surrounding}
  {Vehicles},'' \emph{IEEE Transactions on Intelligent Transportation Systems},
  vol.~19, no.~9, pp. 2965--2976, 2018.

\bibitem{pek2017verifying}
C.~Pek, P.~Zahn, and M.~Althoff, ``Verifying the safety of lane change
  maneuvers of self-driving vehicles based on formalized traffic rules,'' in
  \emph{2017 IEEE Intelligent Vehicles Symposium (IV)}.\hskip 1em plus 0.5em
  minus 0.4em\relax IEEE, 2017, pp. 1477--1483.

\bibitem{tornblom2019abstraction}
J.~T{\"o}rnblom and S.~Nadjm-Tehrani, ``An abstraction-refinement approach to
  formal verification of tree ensembles,'' in \emph{International Conference on
  Computer Safety, Reliability, and Security}.\hskip 1em plus 0.5em minus
  0.4em\relax Springer, 2019, pp. 301--313.

\bibitem{althoff2009model}
M.~Althoff, O.~Stursberg, and M.~Buss, ``Model-based probabilistic collision
  detection in autonomous driving,'' \emph{IEEE Transactions on Intelligent
  Transportation Systems}, vol.~10, no.~2, pp. 299--310, 2009.

\bibitem{bansal2017hamilton}
S.~Bansal, M.~Chen, S.~Herbert, and C.~J. Tomlin, ``{Hamilton-Jacobi}
  reachability: A brief overview and recent advances,'' in \emph{2017 IEEE 56th
  Annual Conference on Decision and Control (CDC)}.\hskip 1em plus 0.5em minus
  0.4em\relax IEEE, 2017, pp. 2242--2253.

\bibitem{fridovich2020confidence}
D.~Fridovich-Keil, A.~Bajcsy, J.~F. Fisac, S.~L. Herbert, S.~Wang, A.~D.
  Dragan, and C.~J. Tomlin, ``Confidence-aware motion prediction for real-time
  collision avoidance,'' \emph{The International Journal of Robotics Research},
  vol.~39, no. 2-3, pp. 250--265, 2020.

\bibitem{9827304}
X.~Wang, Z.~Li, J.~Alonso-Mora, and M.~Wang, ``Prediction-based reachability
  analysis for collision risk assessment on highways,'' in \emph{2022 IEEE
  Intelligent Vehicles Symposium (IV)}, 2022, pp. 504--510.

\bibitem{deo2018convolutional}
N.~Deo and M.~M. Trivedi, ``Convolutional social pooling for vehicle trajectory
  prediction,'' in \emph{Proceedings of the IEEE Conference on Computer Vision
  and Pattern Recognition Workshops}, 2018, pp. 1468--1476.

\bibitem{wang2021probabilistic}
X.~Wang, J.~Alonso-Mora, and M.~Wang, ``Probabilistic risk metric for highway
  driving leveraging multi-modal trajectory predictions,'' \emph{IEEE
  Transactions on Intelligent Transportation Systems}, vol.~23, no.~10, pp.
  19\,399--19\,412, 2022.

\bibitem{zhou2017recurrent}
M.~Zhou, X.~Qu, and X.~Li, ``A recurrent neural network based microscopic car
  following model to predict traffic oscillation,'' \emph{Transportation
  Research Part C: Emerging Technologies}, vol.~84, pp. 245--264, 2017.

\bibitem{li2020prediction}
A.~Li, L.~Sun, W.~Zhan, M.~Tomizuka, and M.~Chen, ``Prediction-based
  reachability for collision avoidance in autonomous driving,'' \emph{arXiv
  preprint arXiv:2011.12406}, 2020.

\bibitem{su2020graph}
J.~Su, P.~A. Beling, R.~Guo, and K.~Han, ``Graph convolution networks for
  probabilistic modeling of driving acceleration,'' in \emph{2020 IEEE 23rd
  International Conference on Intelligent Transportation Systems (ITSC)}.\hskip
  1em plus 0.5em minus 0.4em\relax IEEE, 2020, pp. 1--8.

\bibitem{krajewski2018highd}
R.~Krajewski, J.~Bock, L.~Kloeker, and L.~Eckstein, ``The {highD} dataset: A
  drone dataset of naturalistic vehicle trajectories on german highways for
  validation of highly automated driving systems,'' in \emph{2018 21st
  International Conference on Intelligent Transportation Systems (ITSC)}.\hskip
  1em plus 0.5em minus 0.4em\relax IEEE, 2018, pp. 2118--2125.

\bibitem{turner2018application}
D.~Turner, D.~Andresen, K.~Hutson, and A.~Tygart, ``Application performance on
  the newest processors and {GPUs},'' in \emph{Proceedings of the Practice and
  Experience on Advanced Research Computing}, 2018, pp. 1--7.

\bibitem{goutte2005probabilistic}
C.~Goutte and E.~Gaussier, ``A probabilistic interpretation of precision,
  recall and f-score, with implication for evaluation,'' in \emph{Advances in
  Information Retrieval: 27th European Conference on IR Research, ECIR 2005,
  Santiago de Compostela, Spain, March 21-23, 2005. Proceedings 27}.\hskip 1em
  plus 0.5em minus 0.4em\relax Springer, 2005, pp. 345--359.

\bibitem{mullakkal2020hybrid}
F.~A. Mullakkal-Babu, M.~Wang, B.~van Arem, B.~Shyrokau, and R.~Happee, ``A
  hybrid submicroscopic-microscopic traffic flow simulation framework,''
  \emph{IEEE Transactions on Intelligent Transportation Systems}, vol.~22,
  no.~6, pp. 3430--3443, 2020.

\bibitem{kurtc2020studying}
V.~Kurtc, ``Studying car-following dynamics on the basis of the {highD}
  dataset,'' \emph{Transportation Research Record}, vol. 2674, no.~8, pp.
  813--822, 2020.

\bibitem{mittal2014survey}
S.~Mittal and J.~S. Vetter, ``A survey of methods for analyzing and improving
  gpu energy efficiency,'' \emph{ACM Computing Surveys (CSUR)}, vol.~47, no.~2,
  pp. 1--23, 2014.

\end{thebibliography}
\end{document}